\documentclass[preprint,12pt]{elsarticle}
\usepackage{hyperref}
\usepackage{breakurl}
\usepackage{multirow}
\usepackage[table]{xcolor}

\newcommand{\etal}{\textit{et al}.}
\newcommand{\ie}{\textit{i}.\textit{e}.}
\newcommand{\eg}{\textit{e}.\textit{g}.}
\newcommand{\wrt}{\textit{w}.\textit{r}.\textit{t}.}

\usepackage{gensymb}
\usepackage{amsmath}
\usepackage[capitalise]{cleveref}
\usepackage{tikz}
\def\checkmark{\tikz\fill[scale=0.4](0,.35) -- (.25,0) -- (1,.7) -- (.25,.15) -- cycle;} 

\usepackage{graphicx}
\usepackage{amssymb}

\usepackage{lineno}

\makeatother

\journal{Robotics and Autonomous Systems}

\begin{document}
\begin{frontmatter}


\title{A Benchmark for Point Clouds Registration Algorithms}
\tnotetext[1]{Contributions: \\conceptualization, methodology, software and writing--original draft preparation S.F;\\ writing--review and editing, S.F., D.C., A.L.B., M.V. and D.G.S.;\\ supervision and project administration D.G.S..}


\author{Simone Fontana $^{1}$, Daniele Cattaneo $^{1,2}$, Augusto L. Ballardini$^{1,3}$, Matteo Vaghi$^{1}$ and Domenico G. Sorrenti $^{1}$}

\address{%
$^{1}$Università degli Studi di Milano - Bicocca, Milano, Italy\\
$^{2}$currently with Albert Ludwigs Universität Freiburg, Freiburg im Breisgau, Germany\\
$^{3}$currently with Universidad de Alcala de Henares, Alcala de Henares, Spain}

\begin{abstract}
Point clouds registration is a fundamental step of many point clouds processing pipelines; however, most algorithms are tested on data that are collected ad-hoc and not shared with the research community. These data often cover only a very limited set of use cases; therefore, the results cannot be generalised. 
Public datasets proposed until now, taken individually, cover only a few kinds of environment and mostly a single sensor.
For these reasons, we developed a benchmark, for localization and mapping applications, using multiple publicly available datasets. In this way, we are able to cover many kinds of environment and many kinds of sensor that can produce point clouds. Furthermore, the ground truth has been thoroughly inspected and evaluated to ensure its quality. For some of the datasets, the accuracy of the ground truth measuring system was not reported by the original authors, therefore we estimated it with our own novel method, based on an iterative registration algorithm.
Along with the data, we provide a broad set of registration problems, chosen to cover different types of initial misalignment, various degrees of overlap, and different kinds of registration problems.
Lastly, we propose a metric to measure the performances of registration algorithms: it combines the commonly used rotation and translation errors together, to allow an objective comparison of the alignments.
This work aims at encouraging authors to use a public and shared benchmark, instead of data collected ad-hoc, to ensure objectivity and repeatability, two fundamental characteristics in any scientific field.

\end{abstract}


\end{frontmatter}


\section{Introduction}
Point clouds registration, \ie{}, the alignment of two point clouds, is a very well studied problem, for which many solutions have been proposed. However, most solutions have been tested on very few data and, even worse, usually in specific scenarios, which do not allow for a fair generalisation of the results. Moreover, the data are often collected specifically for a limited set of experiments and not shared with the rest of the community. As a consequence, a fair and objective comparison among different approaches is often impossible. As long as a common benchmark is not available, this problem tends to perpetuate itself: since no common benchmark exists, authors have to collect new data for their experiments.
Indeed a few attempts have been made to design a shared evaluation benchmark. Although worthy, these solutions often cover only a limited set of use cases of point clouds registration algorithms and were collected using a single sensor. Therefore, even though their use is a huge step forward compared to using ad-hoc data, they are not a definitive solution yet.
For these reasons, we propose a benchmarking protocol for point clouds registration algorithms applied to robotic localization and mapping applications. We call it benchmarking protocol because it is composed of not only the data but includes a complete evaluation protocol that authors should use to obtain results that can be compared fairly and objectively. Since many datasets of point clouds are already available, but none covers a large range of use cases and types of sensor, we decided to avoid collecting new data. Instead, we used a set of publicly available sequences from various sources.  However, providing the data is only a part of a benchmark for point clouds registration algorithms. Providing a well-designed set of initial misalignments to apply to the data is as important as the data themselves, because the misalignment, together with the point clouds, defines the actual problem to solve. Despite this, most of the sequences we used did not come with a set of initial misalignments. Therefore, there was no shared way of using the data. Moreover, there was no shared metric to evaluate the results, making a comparison not possible. Finally, part of the sequences we used came with a ground truth whose accuracy was not provided.
The main characteristics of our benchmark are:
\begin{itemize}
\item it covers a quite comprehensive set of use cases, including registrations between clouds coming from different sensors. A scenario very rarely tested in the literature, but common in real life;
\item it allows testing of both global and local registration algorithms;
\item it includes an evaluation protocol that uniformly covers various degrees of overlap and various amounts of misalignment;
\item it includes a metric that reliably combines both the rotation and translation error together into a single measure;
\item it is composed of data whose ground truth has been inspected to ensure its quality and whose accuracy has been estimated;
\item it is freely available and uses only freely available data;
\end{itemize}

\section{Related Work}
There are two big categories of point clouds registration problems: global and local. We have a global registration problem when we align two point clouds without any prior information on their relative pose. On the other hand, the problem is local when we have a prior rough guess on the relative pose, which has to be improved. Algorithms aimed at local registration are usually not effective for global problems, because they often make use of local optimisation techniques and heuristics that could get stuck in local minima. On the other hand, most global algorithms do not provide precise results, thus their solutions usually need a refinement with a local technique. In other words, they are used to estimate the rough initial guess that local registration techniques need. For this reason, they are often called \emph{coarse} registration techniques, while local algorithms are also known as \emph{fine} registration techniques.

Global point clouds registration is usually achieved through the use of geometric features, which are, basically, a representation of salient points of the underlying surface represented by the cloud. 3D features are used in a similar way to what is done with 2D features extracted from images.

Examples of features used for point clouds registration are PFH~\cite{rusu2008aligning}, their faster variant FPFH~\cite{rusu2009fast} and angular-invariant features~\cite{jiang2009registration}. Moreover, Sehgal \etal{} developed an approach, derived from the computer vision world, that uses SIFT features extracted from a 2D image generated from the point cloud~\cite{sehgal2010real}.

These descriptors are usually matched between point clouds using a data association policy based on the closest distance (in the descriptors' space). The matches are then used to estimate a rototranslation, using algorithms such as RANSAC \cite{fischler1981random}. Recently, solutions to the feature extraction problem based on neural networks have been proposed.  Examples are 3dMatch \cite{zeng20173dmatch} and 3DSmoothNet \cite{gojcic2019perfect}. However, the transformation estimation step is still performed using traditional approaches. Instead, Pointnetlk \cite{aoki2019pointnetlk} and Pcrnet \cite{sarode2019pcrnet} are neural network-based solutions that combine both the feature matching and the transformation estimation steps.
 
There are two main drawbacks to feature-based point clouds registration. First of all, it is usually a slow process: detection of keypoints and computation of the descriptors are computationally expensive. Secondly, and most importantly, the resulting alignment is often not very accurate. For these reasons, feature-based registration is mostly used not \emph{per se}, but to estimate an initial guess that will be refined later on with other techniques. This is mainly due to the large number of outliers returned by current feature matching techniques.

Recently, methods aimed at global registration even in presence of many outliers, with execution times comparable to those of local techniques, have been developed. Zhou \etal{} presented a work that significantly improves the results and speed of convergence of global registration techniques in presence of outliers  \cite{zhou2016fast}; however, it has not been proved to outperform the best local methods yet. Another example is the work of Yang \etal{} that developed TEASER++, a fast algorithm for global registration in presence of a large number of outliers or even without correspondences at all \cite{yang2020teaser}.

Local registration techniques, on the other hand, do not usually employ any feature. ICP is the first and most famous member of this category. It was originally developed independently by Besl and McKay~\cite{besl1992method}, Chen and Medioni~\cite{chen1991object}, and Zhang~\cite{zhang1994iterative}. Although its first introduction dates back to 1991, it is still the \emph{de facto} standard for point clouds registration. ICP assumes that the point clouds are already roughly aligned and aims at finding the rigid transformation, \ie{}, a rototranslation, that best refines the alignment. Rather than looking for correspondences using keypoints and descriptors, ICP greedily approximates these correspondences by iteratively looking for the closest point to each point, to improve the alignment at each step.

Many different variants of ICP have been proposed; they usually speed up the algorithm or improve the quality of the result. For an extensive review and comparison of ICP variants, see the work of Pomerlau \etal{}~\cite{pomerleau_comparing_2013}. A variant of ICP that is worth mentioning is Generalized ICP (G-ICP)~\cite{segal_generalized-icp._2009}. G-ICP modifies the standard ICP algorithm by incorporating the covariances into the error function. In this way, it usually obtains better results, but at the expense of computation time. Indeed, the resulting optimisation problem cannot be expressed as a linear system and therefore has to be solved using a generic non-linear optimization algorithm, such as Levenberg-Marquard. Another variant of ICP, specifically aimed at dealing with noise, is Probabilistic Point Clouds Registration (PPCR)~\cite{agamennoni2016point}. It was derived applying statistical inference techniques on a fully probabilistic model. In that proposal, each point in the source point cloud is associated with a set of points in the target point cloud; each association is then weighted so that the weights form a probability distribution. The result is an algorithm that is similar to ICP but more robust \wrt{} noise and outliers.

Local registration algorithms that do not use a nearest-point approximation exist too. For example, NDT~\cite{biber2003normal} represents the point clouds using a set of Gaussians and tries to align them by looking for the most probable alignment. 

Although many different and efficient solutions to the problem of point clouds registration have been proposed, many studies test the proposals only on a few data. Moreover, these data are often collected \emph{ad-hoc}. This approach leads to a severe problem: often there is no direct comparison with other solutions. Even when a comparison is proposed, the exiguity of the data makes it less relevant and not objective. Furthermore, comparing every single solution in a paper is impossible, given the huge number of algorithms in the literature. Therefore, the necessity of shared data and methods to test registration algorithms.

Few meaningful comparisons of registration techniques exist in the literature. Donoso \etal{} compared various ICP variants, including G-ICP, on different outdoor settings~\cite{Donoso2017}, demonstrating that no single algorithm is better than the others in each scenario. However, the datasets used are limited to a single outdoor environment and have been recorded using only a single sensor; therefore, their results cannot be generalized. Cheng \etal{} compiled an extensive review of point clouds registration algorithms~\cite{Cheng2018}. They concluded that a shared and complete dataset, as well as an evaluation system, is necessary, since no real evaluation can be done based on the existing literature. Maiseli \etal{} too pointed out this issue in their review~\cite{Maiseli2017}.
An essential work in this field is that of Pomerleau \etal{}~\cite{Pomerleau2012}, who proposed a benchmark for comparing registration techniques. They collected a series of sequences of point clouds (from now on called the \emph{ETH datasets}) in various environments: indoor structured, outdoor unstructured, and in multiple seasons. Along with the datasets, they also proposed an evaluation protocol. Although it is an important contribution to the literature about point clouds registration, it has some drawbacks, some even recognized by the authors. First of all, the datasets have all been collected using the same sensor. A fair evaluation of a registration algorithm should use datasets produced using many kinds of sensor. Different sensors suffer of different probability distributions of noise, indeed. Another drawback is the use of separate translation and rotation errors to measure the performance of the algorithms. Even though these errors are well defined, having two separate metrics to measure the performance is unpractical. In case an algorithm has a lower translation error than another one, but a larger rotation error, there is no way to decide which one is the best one. We think that to obtain a useful comparison, this ambiguity has to be solved. For this reason, we propose the use of a metric that combines both the translation and the rotation errors. The ETH dataset have been used in several comparisons, such as that of Babin \etal{}, who compared many outlier rejection methods for ICP~\cite{Babin2018}, or those of Magnusson \etal{}~\cite{Magnusson2015}, and Petricek \etal{}~\cite{Petricek2017}. Its use proves that a common and ready to use benchmark is a welcomed addition to the literature on point clouds registration. Another important benchmark for point clouds registration is the 3DMatch geometric registration benchmark \cite{zeng20173dmatch}. However, it is mainly aimed at global registration techniques for scene reconstruction, while our main goal is localization and mapping performed either with local or global methods; Moreover, it contains data collected with only a single kind of sensor (a handheld RGBD sensor).

RESSO \cite{chen2017plade} and WHU-TLS \cite{dong2020registration} are other datasets composed of several different sequences for point clouds registration. While they have some goals in common with the work we propose, such as covering a wide range of environments, they still have some limitations. First of all, they have been collected using a single kind of sensor. While they show an analysis of the overlap between the point clouds, the different degrees of overlap are not covered uniformly, \ie{} some kinds of problems are more represented than others. Moreover, differently from the ETH and our benchmarks, no set of initial misalignments is proposed. Therefore, they do not allow analysing how an algorithm behaves with different degrees of misalignment, on the same data. Finally, they lack a combined metric to objectively compare the results.
\section{Materials and Methods}
Several reasons led us to propose a new benchmark for point clouds registration. First and most importantly, there are no generally accepted test cases for registration algorithms covering a large range of situations. Therefore, authors proposing novel solutions have to either collect new data themselves or use already available datasets. However, no single existing dataset covers all the use cases and possible scenarios of a registration algorithm.
For this reason, authors have to choose the right datasets among the many available. These datasets will probably be in different formats, with a ground truth of variable reliability and accuracy, posing additional work on the authors to prepare the testing environment.
What happens, in practice, is that most of the works are tested on a single or very few datasets, thus the results are inevitably less generalizable. For example, a registration algorithm that uses geometric features could get excellent results in a structured indoor scenario but could perform much worse in an unstructured outdoor one, where geometric features cannot be extracted reliably. On the other hand, an algorithm could exploit the density in a point cloud to reconstruct a surface to improve the registration, but this could not be possible on sparse point clouds, such as those produced from digital elevation maps. The noise pattern could influence the results of a registration too.
Besides being useful for authors, which will have a ready-to-use testing protocol that also permits fair comparisons with other existing solutions, the proposed benchmark is an essential help in choosing the right algorithm for a specific application. Since the benchmark covers many different situations, once authors will have published results, users will be able to select the best solution for their specific applications, without the need for additional experiment or comparison among the many existing solutions, which would require a lot of time and an in-depth study of the literature.
These are the requirements we considered while designing the benchmark:
\begin{enumerate}
    \item it should be a complete testing protocol, not only a dataset or a group of datasets. It should describe how to perform the experiments and how to measure the results. This is essential, since using the same data in different ways is only slightly better than using ad-hoc data;
	\item it should cover as many different settings as possible. Examples include: indoor structured scenes, outdoor unstructured, outdoor structured, with and without moving objects, large and small scale problems;
	\item the data used should come from many sensors. Different sensors have different noise properties and produce point clouds with different densities. \eg{}, some RGB-D cameras, similarly to stereo rigs, are triangulation-based sensors, therefore the error on the measure of a distance increases quadratically with the distance~\cite{matthies1987error}. This does not happen with time-of-flight sensors, such as LiDARs and other RGB-D cameras, whose error on the distance is constant. On the other hand, the density of a point cloud produced with a LiDAR can be much lower than that of point clouds produced with an RGB-D camera, given the much larger range. For these reasons, using data coming only from a single sensor leads to results that cannot be generalized;
	\item it should cover registration problems with both large and small overlap between the two point clouds. An algorithm could perform very well with large overlap problems, but fail when the overlap is smaller. On the contrary, an algorithm performing very well with low overlaps could be outperformed on easier problems;
	\item the benchmark should be useful for testing both local and global registration algorithms; therefore, it should be composed of problems with various degrees of misalignment (initial perturbation);
	\item the benchmarking protocol should include a single metric to compare two algorithms objectively. This sorts out the use of separated translation and rotation errors, as done so far in most of the literature in the field;
	\item the benchmark should include a reliable ground truth, whose accuracy should be provided;
	\item it should be freely available. Therefore, it should include only freely available data.
\end{enumerate}

\subsection{Initial perturbation}
Two factors influence greatly how challenging a registration problem is: the amount of overlap between the point clouds and how far the source point cloud is from the final pose, \ie{}, the initial perturbation or misalignment. With initial perturbation we mean the displacement of the source point cloud \wrt{} its ground truth pose. That is, the rototranslation that a registration algorithm should estimate. The initial perturbation is a fundamental characteristic of a registration problem: the larger it is, the harder the problem becomes.

To test different levels of perturbation, we used the following protocol. For local registration problems, we selected the boundaries of the set of possible transformations, that is, the maximum and minimum magnitudes of the rotation and translation. While for the rotations we could use the same boundaries for every sequence ($0$ and $30$ degree), those of the translations are dataset specific. The effect of a translation on a point cloud, indeed, is dependent on the scale of the cloud: a translation of $1$ meter of a cloud representing an object could be a very hard problem, while on a point cloud representing a city could be considered an easy problem. The actual values have been chosen taking into consideration the accuracy of the ground truth and are available on our repository (\url{https://github.com/iralabdisco/point\_clouds\_registration\_benchmark}). Moreover, for the local registration problems, while choosing the boundaries of the initial transformations' set, we also considered the typical accuracy of systems used to provide an initial guess to point clouds registration algorithms.

For each pair of point clouds, we randomly sampled a set of initial perturbations, each composed of a rotation combined with a translation. The rotation and the translation have been sampled separately although with the same technique. First, we uniformly sampled an axis, which represents either the direction of the translation or the rotation axis. Then, we sampled a magnitude from a uniform distribution with the appropriate boundaries.  The cardinality of the sets of initial perturbations has been carefully tuned to ensure an adequate coverage of the space, but, at the same time, not to require a huge number of tests, which would discourage authors from using the benchmark.

This protocol ensures that our benchmark is not biased towards easier or harder problems, but, instead, covers all the different levels of initial perturbation to highlight strong and weak points of registration algorithms.

A global point clouds registration algorithm should be independent of the initial perturbation. For this reason, for this kind of problem, we sampled only very large transformations. The protocol is the same used for local registration problems, but the boundaries of the uniform distribution are different ($45$ and $180$ degree for the rotation and, again, larger and dataset-specific boundaries for the translation).

\subsection{Overlap}
The point clouds to align do not necessarily represent exactly the same area. Instead, some parts of the scene may be present in a point cloud and not in the other one. The overlap is the part of the scene observed in both point clouds. Usually, the more significant the overlap, the easier the registration problem is. However, an algorithm may behave differently with different levels of overlap. For this reason, we also tested various degrees of overlap.

We calculated the degree of overlap as the percentage of points in a point cloud having a correspondent in the other point cloud (aligned using the ground truth). Since most points will not have an exact correspondent, two points form a correspondence if their distance is lower than a dataset-specific threshold. Since this threshold influences the result, the overlap cannot be compared among different datasets. That is, we can say that two point clouds have a lower/higher overlap with respect to another pair only if they come from the same dataset, since their overlaps have been calculated using the same threshold. This is exactly how we use this measure; therefore, it is perfectly acceptable for our goals.

We think that, to keep the comparison fair, different degrees of overlap should be uniformly represented in each sequence, because the overlap is one of the characteristics that define the difficulty of a registration problem. For this reason, we used the following algorithm to ensure uniform coverage:
\begin{enumerate}
	\item we calculate the overlap of each possible pair of point clouds and discarded the pairs with less than $40\%$ of overlap for local problems and $60\%$ for global ones. These values have been chosen to challenge registration algorithm but, at the same time, to represent the real usage conditions in robotic applications;
	\item we divide the range of overlaps into ten intervals of equal size;
	from each interval we randomly choose ten pairs of point clouds;
	\item if an interval has less than ten members, the remainder are chosen randomly from the whole set of pairs of point clouds.
\end{enumerate}

\begin{figure}
	\centering
		\includegraphics[width=1\linewidth]{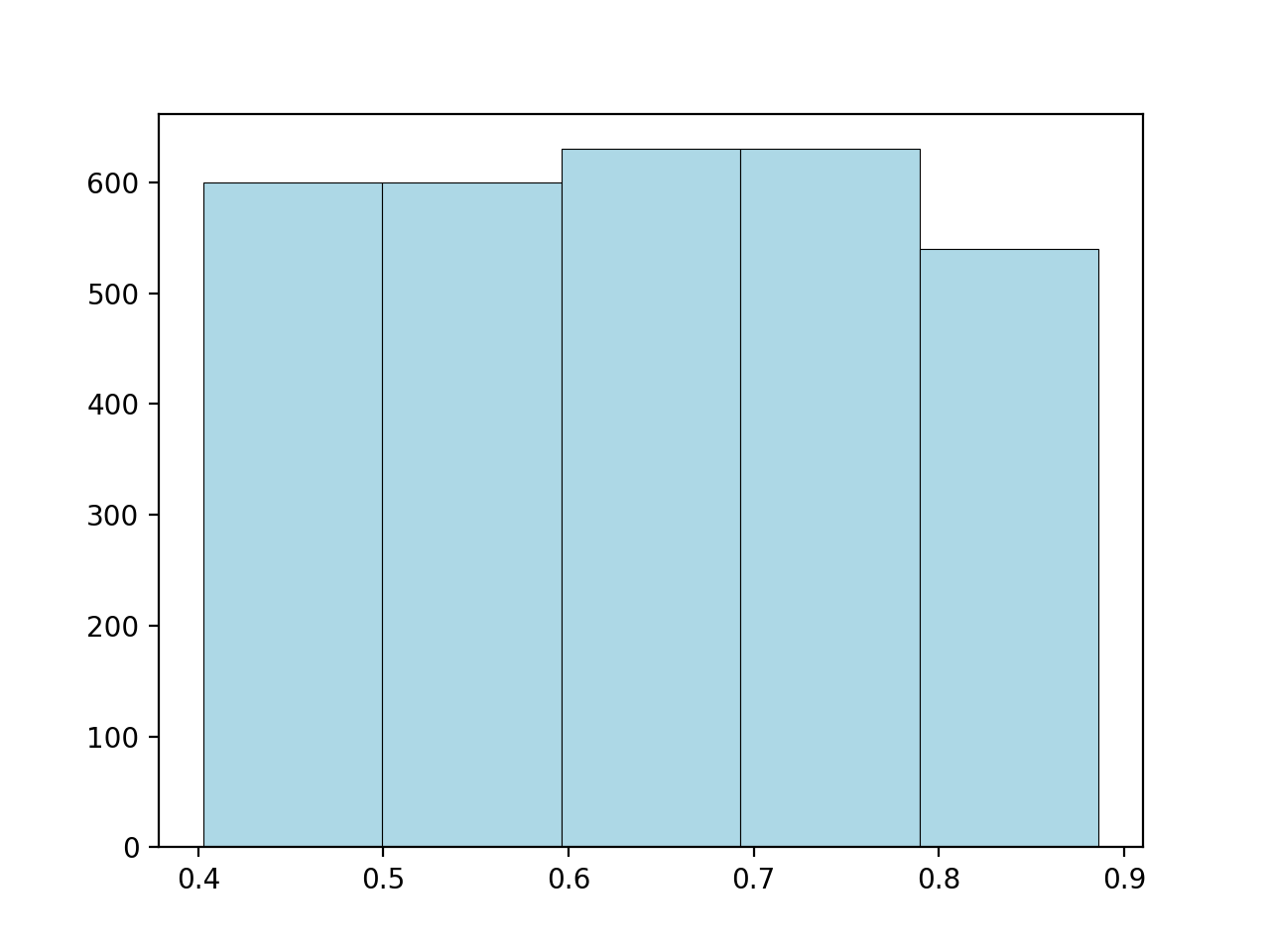}
	\caption{A histogram showing the overlaps of the registration problems that have been selected from the ETH-Apartment sequence. On the x-axis we have the degree of overlap, on the y-axis the number of elements in that range of overlap. Bins are almost equal in size. The differences are negligible and due to our approximate algorithm and to the relatively low sample size. Histograms for other sequences are available on the project web page: \url{https://github.com/iralabdisco/point_clouds_registration_benchmark}	\label{fig:overlap_eth}}
\end{figure}

While this algorithm does not guarantee an exact uniform coverage, in practice, it works reasonably well, as can be seen in \cref{fig:overlap_eth}. 
As an alternative to the described protocol, we could align each point cloud in a given sequence with any other point cloud, with an overlap above a threshold, in the same sequence. We discarded this option for two reasons. First of all, the number of registration problems would become too high, discouraging authors from using the benchmark. Secondly, this protocol would not ensure a uniform coverage of the various degrees of overlap, a fundamental characteristic for a registration benchmark, in our opinion.

For the ETH datasets \cite{Pomerleau2012}, the authors already provide a set of registration problems, \ie{} pairs of point clouds with an associated initial perturbation. Nevertheless, we decided to use registration problems generated with our algorithms, rather than the originals, for two reasons. First, we want to use the same procedure for each dataset. Second, the set of problems provided with the ETH datasets is too large to be used in conjunction with other datasets. Too many  experiments, in our opinion, would discourage researchers from using the benchmark. As reference, our benchmark requires 3000 experiments per sequence for testing local registration algorithms, while the original ETH protocol requires 64000 experiments per sequence. Considering that our benchmark is composed of many more datasets, the number of registration problems would become too large. The time requirement to run the Probabilistic Point Clouds Registration algorithm \cite{agamennoni2016point} on the benchmark is about $14$ hours on a $2.2$GHz Intel Core i7 with $16$GB DDR3 RAM and a SSD hard drive, while this time largely depends on the hardware used and on the parameters and the algorithm tested, we think that further increasing this time would discourage authors from using the benchmark.

Summarizing, for each sequence, we sampled a list of pairs of point clouds, each corresponding to a registration problem, ensuring uniform coverage of the different levels of overlap. For each pair, we randomly sampled a list of transformations, to ensure a uniform coverage of the different magnitudes of initial misplacement. Therefore, each chosen pair of point clouds is tested with different levels of initial misplacement; in this way, the benchmark is able to highlight how an algorithm behaves with different transformations on the same problem.

\subsection{Error metric}

Choosing the right metric is an essential step in designing an evaluation protocol. Most state-of-the-art works about point clouds registration measure their performances by calculating the distance between the estimated and the ground truth translations and rotations separately, therefore obtaining a rotation error and a translation error. Although correct, this approach does not allow an easy comparison among different results, because it does not produce a single measure. If an algorithm gets a lower error on the translation, but a larger one on the rotation, with respect to another algorithm, there is no objective way of deciding which one is the best. From the comparison perspective, this is unpleasant, since the main goal of a benchmark is to compare results; therefore, we think that using a single metric, which does not combine the two above with an arbitrary weighting parameter, is more appropriate.

The result of a point clouds registration algorithm is a pose, \ie{}, a rotation and a translation. Thus, comparing a result \wrt{} a ground truth means calculating the distance between two poses. This problem is very common in many applications, such as path planning and position precision evaluation. However, the research community has not found a widely accepted solution yet.

Among the many existing solutions, an interesting one is that of Mazzotti \etal{}~\cite{Mazzotti2016}, who use a so-called platonic solid attached to a rigid body to measure the distance between two poses. The actual distance is then calculated using the root mean squared distance between the homologous vertices of the solid. Suggestions on which solid to use are given in their work. This solution could be appropriate for our goals: it allows objective comparisons between results, it is very easy to calculate and has an intrinsic physical meaning (since it is just a mean of Euclidean distances). However, it has a very relevant drawback: two very important parameters affect the metric. Both the size of the solid, but not the type, and where it is placed on the rigid body affect the contribute to the distance of the rotation \wrt{} the translation. Suppose that the solid is placed at the origin of the reference frame of the two point clouds; using a very large solid will give more importance to the rotation, since points far from the origin will be displaced by a higher distance, if the solid is rotated, than if they were closer to the origin. Placing the solid in  different position than the origin of the reference frames has the same effect: placing it further gives more importance to the rotation component of the rototranslation.
Similar problems, with parameters that would bias the metric, also arise with the work of Di Gregorio~\cite{DiGregorio2008}, who proposed a method to generate a family of metrics. The parameters are tuned so to constrain the maximum displacement of the rigid body, a feature useful in path planning, but not appropriate for evaluating point clouds registration algorithms.

Inspired by the work of Mazzotti \etal{}~\cite{Mazzotti2016}, we propose a new metric. Rather than using the vertices of a platonic solid, we used the points of the actual point cloud. Therefore, the proposed metric is calculated using the root mean squared distance between homologous points of the source point cloud, after the execution of an algorithm, and the same point cloud at the ground truth pose. Since the point cloud at the ground-truth and the one aligned with an algorithm are actually the same point cloud, although displaced, associating a point in the former with the homologous in the latter is trivial.
To make the metric scale-invariant, each distance between homologous points is divided by the distance of that point \wrt{} the centroid of the point cloud, \ie{} the mean of the points. Without this last step, the metric would depend on the size of the point cloud: the same rototranslation applied to a larger point cloud would result in a higher error than if the point cloud was smaller. Consequently, the results of an algorithm on different pairs of point clouds would not be comparable and statistics such as the mean or the median of the performance on the different registration problems would be meaningless. For these reasons, this last step is fundamental in making the metric useful for comparisons. This is an important difference \wrt{} other measures based on the distance between homologous points: meaningful statistics are essential when evaluating an algorithm on such a large set of problems.

Given the same point cloud in different poses $P$ and $G$, of cardinality $n$ and with point $p_i$ in $P$ corresponding to $g_i$ in $G$, with $0\leq i \leq n$ and $\overline{p}$ being the centroid of $P$ , the distance between $P$ and $G$ is
\begin{equation}
    \delta(P,G) = \frac{\sum_{i=0}^{n} \frac{\|p_i - g_i\|_2}{\|p_i-\overline{p}\|_2}}{n}
\end{equation}

where $\|\mathbf{x}\|_2$ is the Euclidean norm of vector $\mathbf{x}$.

To be a well-formed distance, the proposed metric should satisfy three constraints:
\begin{enumerate}
    \item symmetry, \ie{}, $\delta(P,G) = \delta(G,P)$;
    \item positive definiteness, \ie{}, $\delta(P,G) > 0$ if $P \neq G$ and $\delta(P,G) = 0$ if $P=G$;
    \item triangle inequality, \ie{}, $\delta(P,G) \leq \delta(P,H) + \delta(H,G)$.
\end{enumerate}
These three requirements are easily verified since the proposed metric is an average of Euclidean distances that satisfy the constraints. The sum is a symmetric and positive definite operation that complies with the triangle inequality as long as the operands are not negative (this is true in our case since a Euclidean distance cannot be negative).

Besides being a well-formed distance metric, our proposal also satisfies the requirements of our benchmark. It combines the rotation error and translation error in a single value, allowing an immediate and objective comparison among results. Differently from the use of the vertices of a platonic solid, there is no parameter to tune that influences how the metric behaves. By increasing the size of the solid we would be able to give a larger importance to the rotational component; on the other hand, by using the points of the point cloud directly, this parameter is implicit. If there are many points far from the origin, then an error on the rotational component will have a more significant impact on the metric. This is a desirable behaviour since in such cases the rotation would have a greater effect on the registration result.

The only drawback of our proposal is that it requires to iterate through the whole point cloud, instead of through a few vertices. However, this issue is negligible, since the comparison with the ground truth is done only for benchmarking purposes and is not needed online in a real application.

\section{The Datasets}
No single dataset complies with all the requirements we formulated. For this reason, taking advantage of the large number of publicly available point clouds, we decided to base our benchmark on multiple public datasets.

Among the many available, we concentrated on those more relevant to applications like localization and mapping. Therefore, we preferred datasets with sequences representing large and complex environments. Instead, we discarded those more aimed at object or scene reconstruction. The latter is a task often accomplished in a controlled environment, where the poses of the sensor can be measured very accurately. On the contrary, localization and mapping is usually performed on platforms moving in a very dynamic environment, where the poses of the sensors are measured with a large uncertainty; however, the accuracy and precision required are usually lower. Moreover, these problems are often solved with different techniques and using point clouds acquired with different sensors. Also, the overlap between two point clouds may differ according to the application. In object reconstruction, for example, it is usually possible to acquire several point clouds from different points of view, resulting in a great overlap between acquisitions. On the contrary, this is not usually feasible in robotics applications. For these reasons, we think that so different applications require different benchmarks.

\subsection{Ground Truth Evaluation}
Having a reliable and accurate ground truth is a strong requirement that, unfortunately, brought us to discard many datasets that would have been suitable otherwise. We inspected several publicly available datasets and discovered that a surprisingly high number had a very inaccurate ground truth. Moreover, we discovered that often the accuracy reported by the authors does not correspond to the real accuracy of the ground truth. We believe that having a ready-to-use set of sequences whose ground truth has been inspected is one of the major advantages of the proposed benchmark.

A measure of the accuracy of the ground truth is therefore necessary, since it is also the lower limit beneath which the accuracy of an algorithm cannot be evaluated. 

While for some of the datasets the authors provide the accuracy of their ground truth measurement system, this information is not available for others. Since we think that it is an essential part of a benchmark, we decided to evaluate the accuracy of all the datasets with another technique, regardless of whether it had already been reported. We decided to re-evaluate all the accuracies to ensure that every dataset was evaluated in the same way.

To measure the accuracy of the ground truth of a sequence, we tried to align each pair of point clouds, that is, each registration problem, with the Probabilistic Point Clouds Registration algorithm~\cite{agamennoni2016point}, using as maximum distance between associated points (the radius parameter) the value used to calculate the overlap in the corresponding sequence. We used this algorithm because it provides better results than other state-of-the art algorithms, such as NDT or ICP~\cite{agamennoni2016point}, although at the expense of computational time, which is not relevant when estimating the accuracy of the ground truth. As any point clouds registration algorithm, it estimates the rigid transformation between two overlapping point clouds, therefore providing a measure of how misaligned the two clouds are. When applied to two already aligned point clouds, this measure is, essentially, the accuracy of the ground truth.  Indeed, the resulting rototranslation represents how much the algorithm was able to improve the alignment. One drawback of our technique is that the used algorithm, similarly to many other algorithms in this field, has no guarantee of convergence to the globally optimal solution. However, this drawback is mitigated by several factors. First of all, if the two point clouds are already well aligned, such as in the case of alignment starting from the ground truth pose, closest-point based registration algorithms are almost always able to converge to the right solution. Moreover, whilst the closest-point based data association could lead to wrong solutions, the Probabilistic Point Clouds Registration algorithm is guaranteed to converge to the right solution if the right data association is contained among the associations used (under the t-distribution assumption and a proper outlier rejection). Finally, since the algorithm could, nevertheless, give wrong results sporadically, we decided to consider as outliers (therefore not considering them in the evaluation of the accuracy of the ground truth) misalignments with a robust z-score greater than $3.5$. Besides detecting these few outliers, we also inspected them manually and we can confirm that they correspond to errors of the registration algorithm and not to a low accuracy of the ground truth. Therefore, we did not use them for the evaluation of the accuracy of the ground truth.
As a measure of the accuracy of the ground truth of a sequence, we have taken the mean and the standard deviation of the misalignments, estimated by our approach, calculated using the metric proposed for this benchmark, without the scale-invariant normalization (since we need absolute values for the ground-truth evaluation), and excluding the outliers, as mentioned before.

It is important to note that the method we used is not an exact measure of the accuracy of the ground truth, since it is based on a heuristic. Instead, the reported value has to be considered an upper bound to the value of the accuracy of the ground truth. While this solution is not optimal, it is, nevertheless, the best that can be done using the data alone. An exact measure of the ground truth’s accuracy can be performed only with a direct analysis of the ground truth measuring system, as it has been done by Pomerleau \etal{} \cite{Pomerleau2012}. Unfortunately, this is not possible when using existing datasets.
Our method reports a value for the accuracy of the ground truth that is usually greater than the one reported by the original authors (when available). The reason is that, while the methods employed by the original authors measure only the errors introduced by the ground truth measuring system, our method is also influenced by the noise in the point clouds, since our evaluation uses the actual data. 

The accuracies evaluated with our method are reported in \cref{tab:accuracy}. As an example, Figure 2 depicts the ground truth evaluation for the TUM dataset. On the $x$ axis we have the pairs of point clouds used by the benchmark, on the $y$ axis the accuracy of the ground truth. The red line is the mean of the ground truth accuracy, while the green area represents $\pm$ the standard deviation. The full data used for the evaluation of every dataset, along with the corresponding plots, are available on our repository (\url{https://github.com/iralabdisco/point_clouds_registration_benchmark)}.

\begin{table}[]
\centering
\small
{\rowcolors{2}{gray!5}{gray!20}
\begin{tabular}{|l|| c |c|}
\hline
\textbf{Name}                         & \textbf{Mean Error {[}m{]}} & \textbf{Std. Deviation {[}m{]}}\\
\hline
\hline
ETH Dataset                  & 0.05               & 0.02                  \\
Canadian Planetary Emulation & 0.13               & 0.05                  \\
TUM Dataset                  & 0.11               & 0.09                  \\
KAIST Dataset                & 0.04               & 0.03 \\
\hline
\end{tabular}}
\caption{Upper bounds on the accuracy of the ground truth of the datasets. For the datasets composed of multiple sequences, we report the lowest accuracy. The data are available on our repository.\label{tab:accuracy}}
\end{table}

\begin{figure}
	\centering
			\includegraphics[width= 0.7 \linewidth]{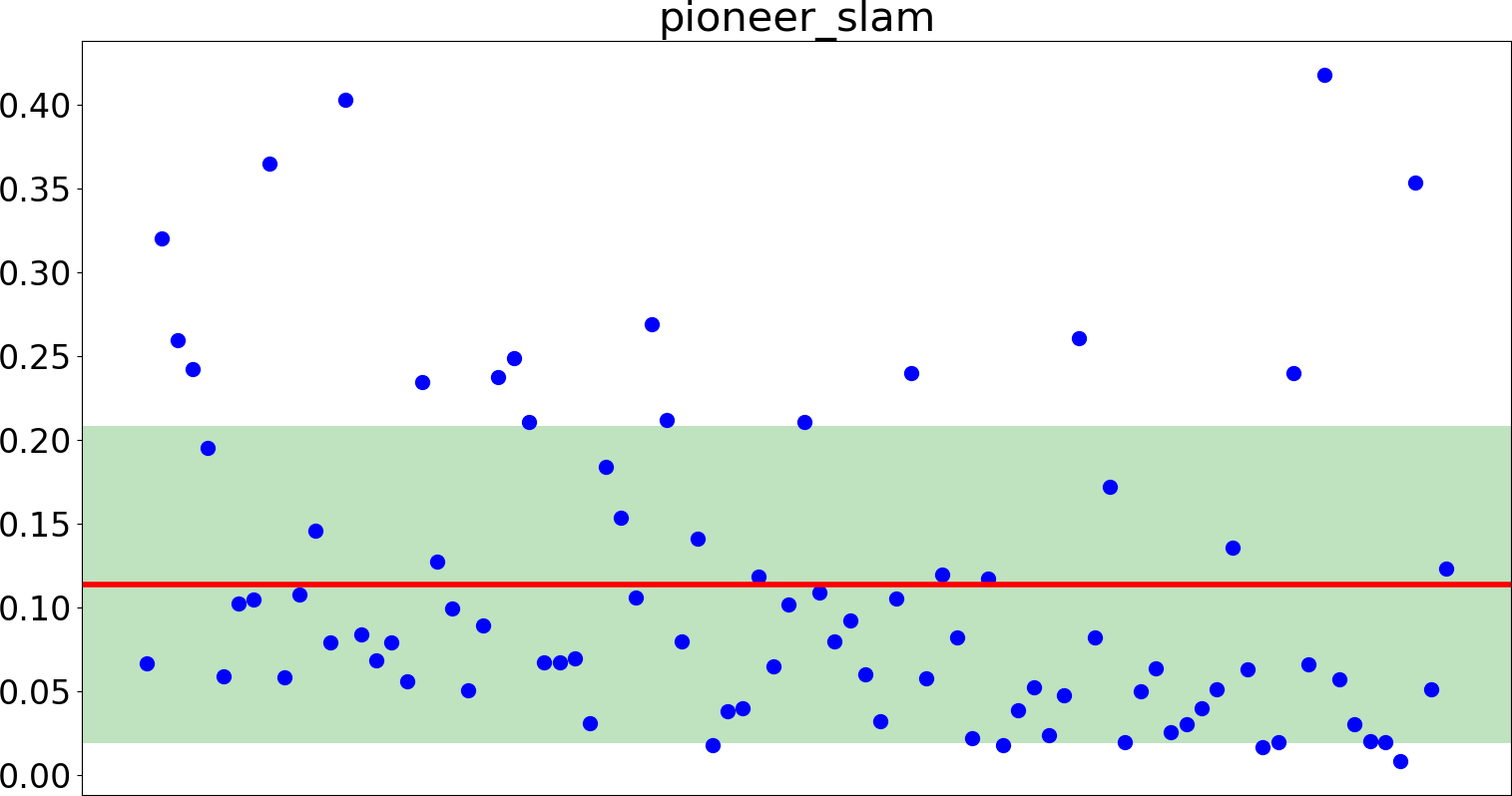}
	\par\bigskip
			\includegraphics[width=0.7\linewidth]{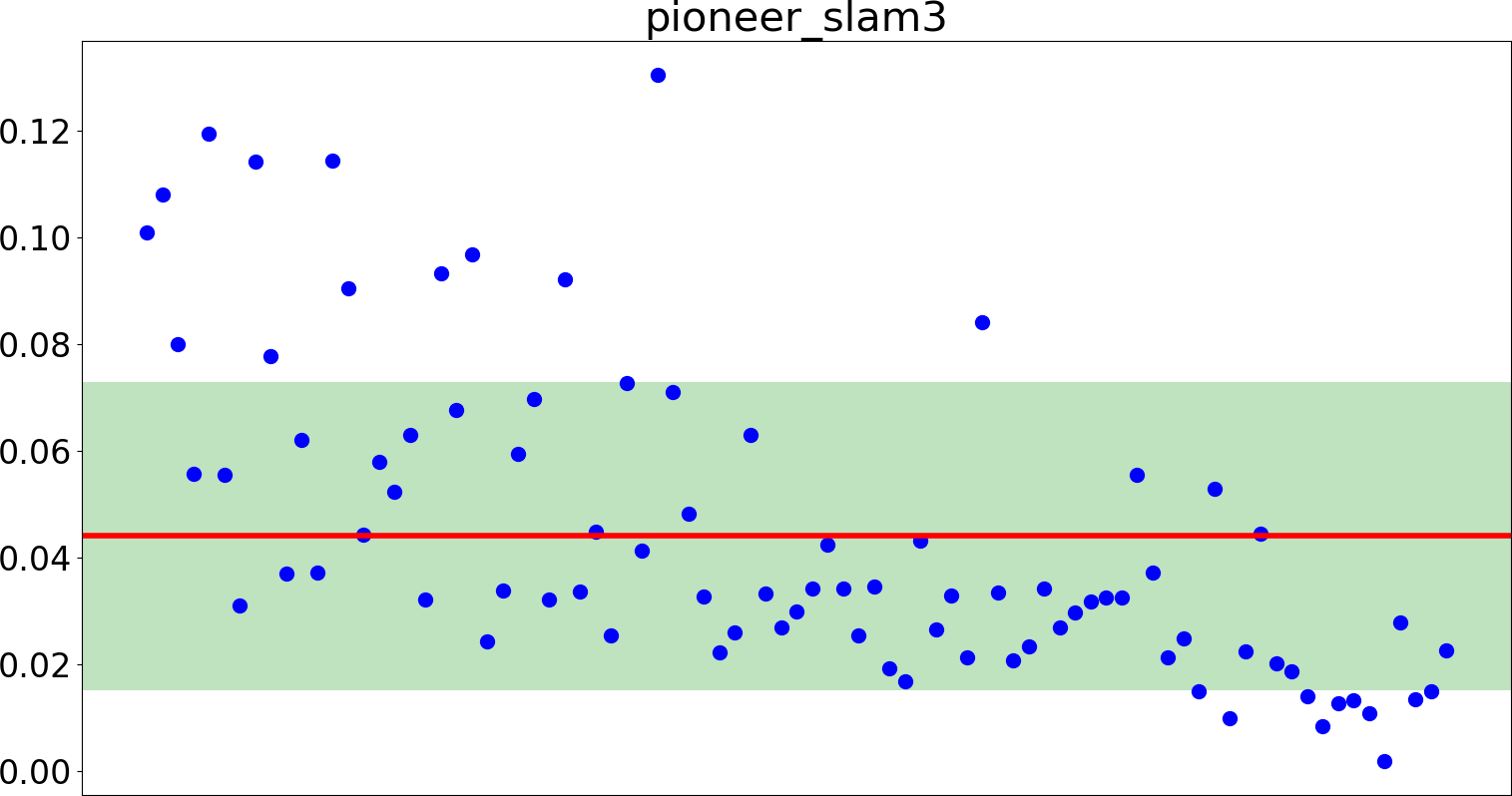}
	\par\bigskip
			\includegraphics[width=0.7\linewidth]{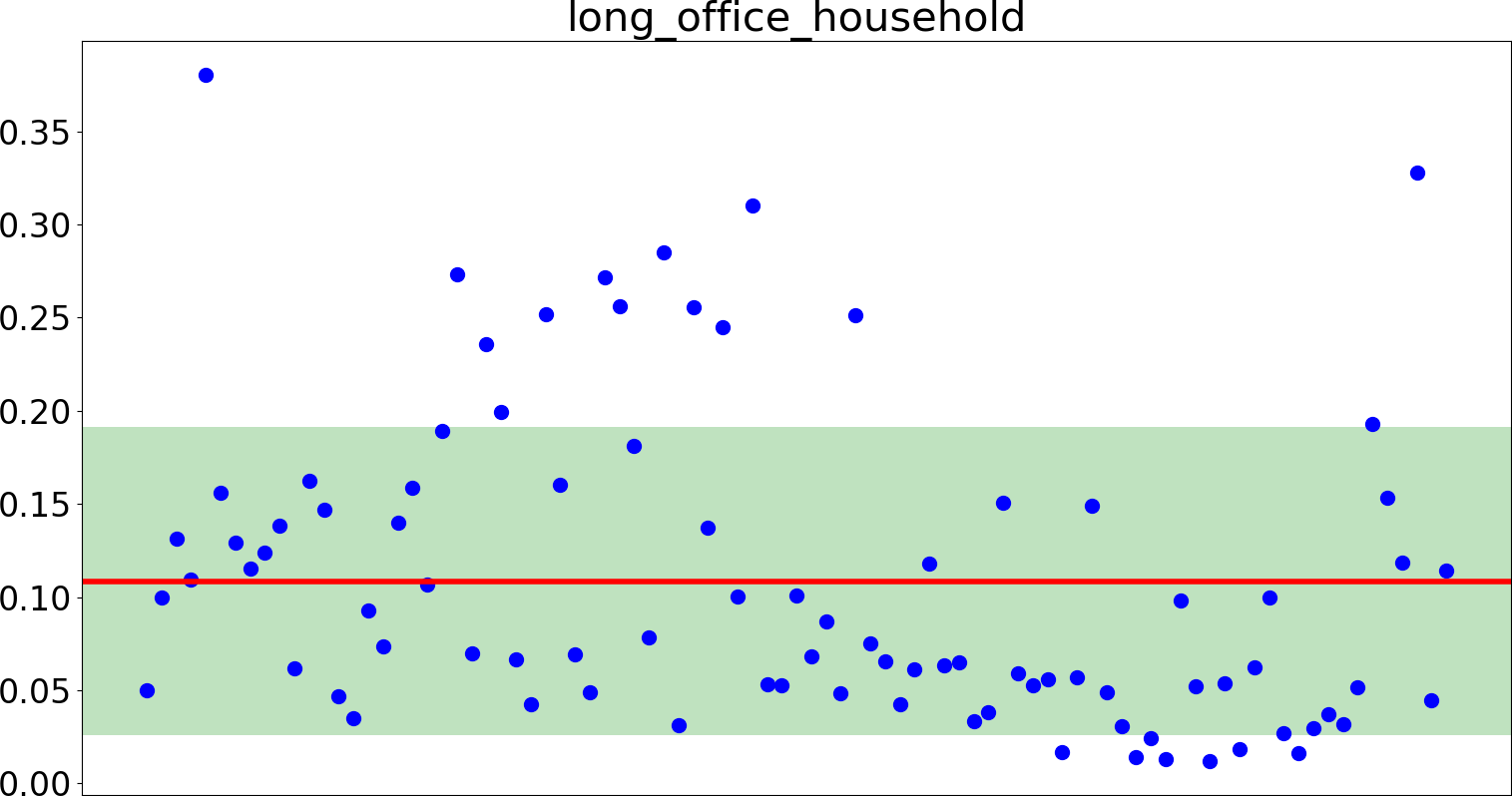}
	\caption{ The result of the evaluation of the ground truth of the TUM datasets. On the $x$ axis we have the pairs of point clouds used by the benchmark, on the $y$ axis the accuracy of the ground truth. The red line is the mean of the ground truth accuracy, while the green area represents $\pm$ the standard deviation.\label{fig:tum_accuracy}}
\end{figure}

The datasets we used are described in the following sections.
\subsection{The ETH Dataset}
We used the dataset presented by Pomerleau \etal{}~\cite{Pomerleau2012}, because it covers a broad set of use cases for registration algorithms. It contains two indoor scenes (apartment and stairs), five outdoor scenes (gazebo\_summer, gazebo\_winter, plain, wood\_summer and wood\_autumn) and a mixed one (hauptgebaude). It includes both structured and unstructured environments, and the indoor ones are not entirely static (there are walking people or furniture moved between scans).
The datasets have been recorded with a Hokuyo UTM-30LX scanning rangefinder. The corresponding ground truth has been measured by the authors with a Leica TS15 base-station, obtaining an accuracy of 1.8mm for the translation and 0.006rad for the rotation. For a complete description of the methodology used, please consult the corresponding paper. The evaluation with our own method is reported in Table 1.
\subsection{Canadian Planetary Emulation Terrain 3D Mapping Datasets}
\begin{figure}
	\centering
		\includegraphics[width=0.9\linewidth]{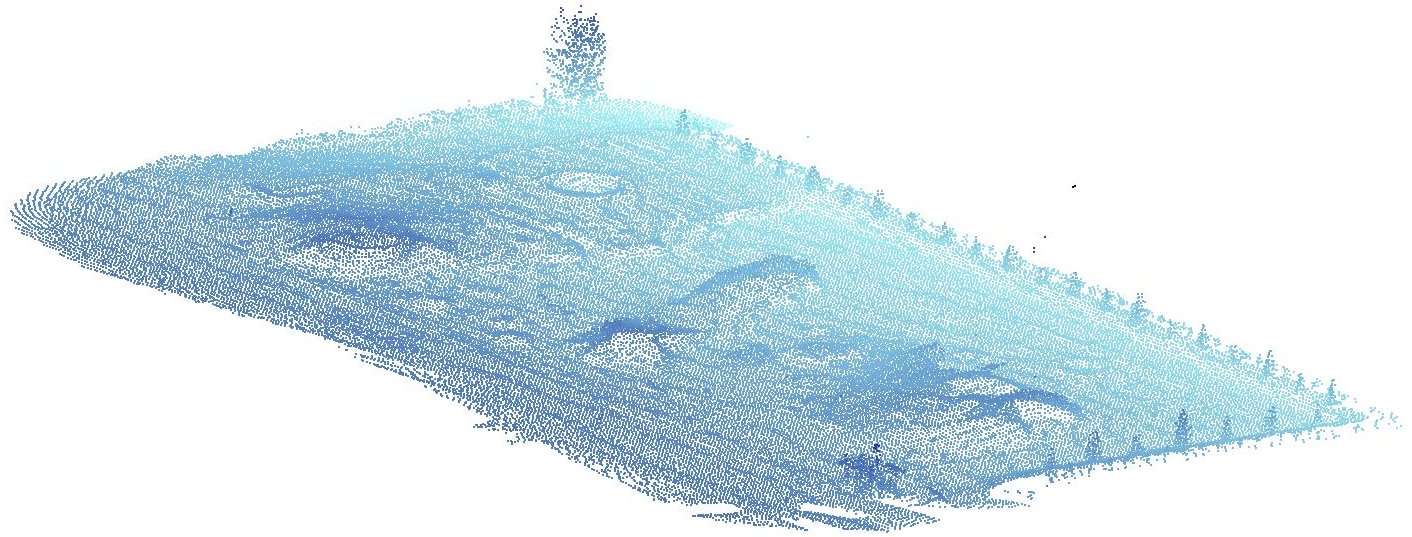}
	\caption{ The map built using the box\_met sequence.\label{fig:box_map}}
\end{figure}

The second dataset is aimed at emulating planetary explorations~\cite{planetary}. We think that this dataset is particularly suitable for testing the performance of registration algorithms in unstructured outdoor scenarios, a setting that poses hard challenges and is often neglected in the literature. The environment is mainly composed of sand, scattered rocks, and some trees at the borders. The whole area has a dimension of $120\times60$ meters.
The almost complete lack of structure is what brought us to choose this dataset: we wanted to test the performance of registration algorithms in one of the hardest scenarios they can encounter \cref{fig:box_map}. Algorithms exploiting geometric features will probably struggle with this dataset. However, the goal of our benchmark is to highlight the strong and the weak points of the various registration techniques, therefore the choice of such a challenging scenario. This dataset represents the typical operating environment of outdoor robots, such as search and rescue, space or agricultural robots, applications often neglected by other datasets. These applications are gaining increasing popularity, especially agricultural robots, and therefore, in our opinion, cannot be neglected by benchmarks anymore.
The dataset is composed of many sequences, acquired in three different facilities, using a Sick LMS291-S05 or a Sick LMS111-10100 laser rangefinder, two popular sensors, mounted on three different robotic platforms. Unfortunately, we could use only two sequences, named \emph{p2at\_met} and \emph{box\_met}, because the others do not have a reliable ground truth. Our manual inspection of the sequences, indeed, revealed several errors in the ground truth.
Since the two sequences depict the same environment, we could add a particular test case to our benchmark, \ie{}, to measure the performance of registration algorithms while aligning a point cloud with a map produced at a different time using a different sensor. Therefore, we built a map with the point clouds acquired with the box platform (from the box\_met sequence) and subsampled it using a voxel grid of fixed size, to simulate a Digital Elevation Map (\cref{fig:box_map}). This map can then be used to localize the other platform by aligning, \wrt{} this map, the point clouds from the p2at\_met sequence.

\begin{figure}
	\centering
		\includegraphics[width=0.9\linewidth]{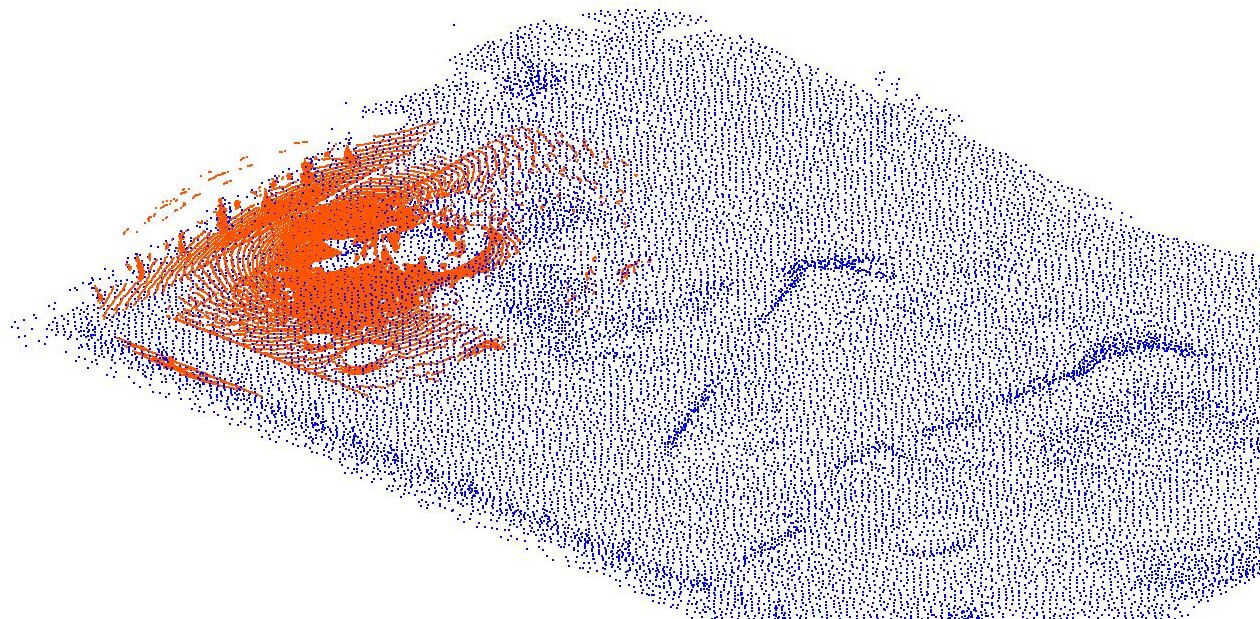}
	\caption{A point cloud from the p2at\_met sequence aligned with the map built using the box\_met sequence, from the Canadian Planetary Emulation dataset. The density of the map is much lower than that of the point cloud.\label{fig:p2at2box}}
\end{figure}

This is an example of registration between two different types of point clouds. Besides the sensor used, the point clouds also differ in their density: the map has a much lower density indeed (\cref{fig:p2at2box}). Thus, this is also an example of dense to sparse point clouds registration. This kind of problem has been long understudied, but is gaining importance as is nevertheless very relevant for real-world robotics applications. As an example, localizing on a map produced with a different sensor, producing sparser point clouds, is a very common use case in robotics. In a word where 3D maps of various environments will be readily available (this is already happening for cities and will likely happen also for other public settings), this use case will be very common. Moreover, for many outdoor areas, digital elevation maps (DEM) are already available. After converting them into point clouds, they can be used for localization tasks. If localization systems want to use 3D maps that are, or will soon be, publicly available, they will have to solve point clouds registration problems between point clouds with different characteristics. These maps, indeed, will likely have a lower density than the clouds produced with an on-board sensor, since they have to represent very large areas. Moreover, these maps will likely be produced with different sensors or completely different techniques. For example, a common technique for producing maps of large outdoor areas is photogrammetry. For these reasons, testing the performances of registration algorithms with point clouds with heterogeneous characteristics should be a fundamental part of a benchmarking protocol.
The ground truth of the chosen sequences has been produced using a differential GPS and an on-board IMU. The authors do not report the accuracy of their ground truth measuring system. Our evaluation is reported in \cref{tab:accuracy}.

\subsection{The TUM Vision Ground RGBD Datasets}
The third dataset we selected is the RGBD dataset acquired by the TUM Vision Group~\cite{sturm12iros}. From the large number of available sequences, we have chosen the most relevant for localization and mapping, as opposed to scene or object reconstruction and classification. These sequences are usually longer and involve navigating through a large and complex scene, rather than turning in place or moving around a single object. Note that we had to discard some sequences because the complete ground truth was not available.
The chosen sequences are:
\begin{itemize}
    \item freiburg3\_long\_office\_household
	\item freiburg2\_pioneer\_slam
	\item freiburg2\_pioneer\_slam3
\end{itemize}
The sequences have been recorded using either a Kinect1 or an Asus XTion, two popular RGBD sensors based on triangulation. Triangulation-based sensors are affected by noise with a different noise pattern than time-of-flight sensors~\cite{matthies1987error}; therefore, experiments performed with the latters cannot be generalised to the formers. Given the increasing availability and reduction in the cost of RGBD triangulation-based sensors and their consequent widespread use, we think that including datasets produced with them is an essential part of any point clouds registration benchmark.

The ground truth has been produced using a very accurate eight-camera motion capture system, resulting in an error on the poses lower than $0.10m$ and $0,009rad$. The evaluation of the accuracy of the ground truth with our own method is reported in \cref{tab:accuracy}.
\subsection{The KAIST Urban Datasets}
Autonomous driving is a field of robotics that is gaining increasing importance for both the research community and companies, such as Google, Tesla, Daimler, BMW, Toyota, and many others. The typical urban environment in which an autonomous vehicle has to operate has many peculiarities. Examples are the presence of many moving objects (pedestrians or other vehicles), scenes with very repetitive patterns or where geometric features cannot be reliably extracted or matched (\eg{}, tree-lined streets). Despite their importance, these elements are usually neglected by datasets not specifically aimed at autonomous driving. Since experimental autonomous road vehicles are already a reality and are already being tested in the real world, we think that a complete benchmark cannot neglect the typical urban scenario in which autonomous cars have to operate.
The KAIST Urban dataset is a collection of data acquired with sensors mounted on a vehicle, driving in various types of scenarios in Korea~\cite{jeong2019complex}. The point clouds we are going to use for our benchmark have been acquired with two Velodyne VLP-16 LiDARs mounted on the left and right of the top of a car. The ground truth has been produced with an RTK-GPS in conjunction with a SLAM system. Since the point clouds from the two LiDARs and the ground truth poses were not synchronized, we decided to associate to each point cloud from the left LiDAR the closest in time ground truth pose. We associated point clouds from the right LiDAR with those from the left one using the same strategy. We manually checked the differences in time implied by such choice, and found they are in the order of a few milliseconds. Therefore, each point cloud of our benchmark is composed of a point cloud produced with the left LiDAR, merged with its corresponding cloud from the right LiDAR. This is how these point clouds will probably be used in a real application, since the transformation between the left and right LiDAR is known and fixed. Using them in conjunction, instead of aligning them separately, could provide more clues in some situations, \eg{} during sharp turns, to a registration algorithm, since the represented area and the overlap with the map would be larger. Even though we want our benchmark to be challenging, we do not want it to be artificially too hard by unnecessarily removing information that would be available in a real-world application.
Although the dataset is composed of many sequences, we decided to use only the one named ``urban05''. Other sequences, very similar to the chosen one, could have been used; however, we decided to use just one to keep the number of experiments required by our benchmark not too high. We avoided sequences representing only a highway and where the observer was only going straight, because they do not represent the complexity of an urban environment, \ie{}, the common operating environment of an autonomous road vehicle.
The ground truth of this sequence has been acquired with a RTK GPS and refined using a SLAM system. The authors do not report the accuracy of their system; therefore, we evaluated it with our method, the result is reported in \cref{tab:accuracy}.

\subsection{Final remarks on the datasets}
With these four groups of sequences, we aim at covering the large variety of possible uses for point clouds registration algorithms for localization and mapping: indoor, outdoor rural, urban, unstructured, structured, and also between clouds with different characteristics. \Cref{tab:datasets} describes the datasets that compose our benchmark, while \cref{tab:char} summarises the characteristics of the various datasets we used and highlights how no single one meets all requirements.

\begin{table}
\centering
\small
{\rowcolors{2}{gray!5}{gray!20}
\begin{tabular}{|p{2cm}||l|p{8cm}|}
\hline
\textbf{Name} & \textbf{Sensor} & \textbf{Peculiarities} \\ 
\hline
\hline
ETH & LiDAR & Dynamic scenes, different kind of environments\\
Planetary & LiDAR & Completely unstructured, very few features, registration \wrt{} a map\\
TUM & RGBD Camera & \\
Kaist & LiDAR & Autonomous driving\\
\hline
\end{tabular}
\caption{The datasets we used and their characteristics.}
\label{tab:datasets}}
\end{table}

\begin{table}
\centering
\scriptsize
{\rowcolors{3}{gray!5}{gray!20}
\begin{tabular}{|l|c|c|c|c|c|c|c|c|c|}
\hline
\textbf{Name} & \textbf{Outdoor} & \textbf{Indoor} & \textbf{Urban} & \textbf{Heter.} & \textbf{Struct.} & \textbf{Unstruct.} & \textbf{RGBD} & \textbf{LiDAR} & \textbf{Dynamic}\\ 
 & & & \textbf{} & \textbf{registration} & & & & &\\
\hline
\hline
ETH  & \checkmark{} & \checkmark{} & X & X & \checkmark{} & \checkmark{} & X & \checkmark{} & \checkmark{}\\
Canadian & \checkmark{} & X & X &\checkmark{} & X & \checkmark{} & X & \checkmark{} & X\\
TUM  & X & \checkmark{} & X & X &\checkmark{}& X & \checkmark{} & X & X\\
KAIST  & \checkmark{} & X & \checkmark{} & X & \checkmark{} & \checkmark{} & X & \checkmark{} & \checkmark{}\\
\rowcolor{green!10}
Proposal & \checkmark{} & \checkmark{}& \checkmark{}& \checkmark{}& \checkmark{}& \checkmark{}& \checkmark{} & \checkmark{} & \checkmark{}\\
\hline
\end{tabular}}
\caption{The characteristics of the datasets we used. With ``heter. registration'' we mean registration between point clouds produced with different sensors or with different densities. With ``dynamic'' we refer to datasets containing moving objects or objects that have been moved between point clouds.
\label{tab:char}}
\end{table}

\section{Software provided with the benchmark} 

On the project web page, \url{https://github.com/iralabdisco/point_clouds_registration_benchmark}, we provide a set of utilities developed to make the benchmark immediate to use. 

We have chosen pairs of point clouds randomly, to cover various degrees of overlap. These registration pairs are available in the form of a text file, containing the following fields: 
\begin{description}
\item[id] A unique identifier of the registration problem; 
\item[source name] The file name of the source point cloud; 
\item[target name] The file name of the target point cloud; 
\item[overlap] The percentage of overlap between the point clouds; 

t1..t12: The elements of the 4x4 transformation matrix representing the initial misplacement to apply. The last line is implicit, since for a rototranslation in homogeneous coordinates it is always the same; therefore, the matrix is the following: 
$\begin{bmatrix}
     t1 & t2 & t3 & t4  \\
     t5 & t6 & t7 & t8  \\
     t9 & t10 &t11 & t12 \\
     0 & 0 & 0 & 1
\end{bmatrix}$
\end{description}

The transformation matrix represents the initial misplacement of the source point cloud and must be applied before proceeding to solve the problem. That is, it is the variable that a registration algorithm has to estimate. 

There are two files for each sequence: one is relative to local registration problems, the other to global ones.  Transformations for global registration algorithms have been sampled from a uniform distribution with a much larger variability, therefore the need for two different sets of problems. 

Since a pair of point clouds corresponds to several registration problems, we decided to provide the clouds correctly aligned, that is, in their ground truth position. Before solving a problem, the transformation described in the corresponding line of the configuration file has to be applied to the source point cloud. 

We think that having all the data in the same format is an essential part of a benchmark that is usable. Initially, all the datasets came in their own format; therefore, they had to be converted to a common one. We decided to use the ASCII PCD format because it can be used with the most popular point clouds library, PCL~\cite{rusu20113d}. The binary version of the PCD would have been more efficient; however, we preferred the ASCII version because the data are expressed in plain text, in order to maintain compatibility with those not using PCL. Writing a text parser is, indeed, a very trivial task. 

Rather than converting all the datasets to the PCD format and storing them on our website, we decided to write a script that downloads the point clouds from the original sources and converts them locally. We think that this choice is more respectful of the original authors since we did not produce the data. Exceptions to this rule are the digital elevation map we built for the Planetary Emulation Dataset, the Kaist datasets that would require a manual registration to the authors’ website and the TUM datasets that would require a complex setup. However, both the Kaist and the TUM datasets are released under the Creative Common 3.0 License, so we are allowed to redistribute it under the same license. 

Finally, the third utility we provide is a library to calculate our metric. Even though it is quite straightforward to compute, to reduce the effort needed to use our benchmark, we decided to develop C++ and Python libraries, compatible with PCL, for calculating the metric. 

Summarizing, to use our benchmark the following steps are necessary: 
\begin{enumerate}
\item use our script to download the data and prepare the environment; 
\item pick the right set of configuration files, either for global or local point clouds registration. Those describing global registration problems have the \emph{\_global} suffix; 
\item for each line in the chosen configuration files, transform the source point cloud with the corresponding initial transformations; 
\item solve the registration problem with the algorithm to test; 
\item report the results, using the proposed metric. 
\end{enumerate}
For the full user guide of the provided software and data, please consult the project web page at \url{https://github.com/iralabdisco/point_clouds_registration_benchmark}. 

The benchmark, including the proposed metric, is also available in TorchPoints3D, a unified framework for deep learning on point clouds \cite{tp3d}.

\section{Example Usage}
To show how our benchmark can be used to measure the performance and to highlight the peculiarities of point clouds registration techniques, we used it to test some popular algorithms. We performed tests using both the set of local registration problems and the set of global problems. For the local registration problems we used: ICP \cite{besl1992method}, one of its best variants, G-ICP \cite{segal2009generalized}, Normal Distributions Transform (NDT) \cite{biber2003normal}, and the Probabilistic Point Clouds Registration algorithm (PPCR) \cite{agamennoni2016point}. For the global registration problems we used: TEASER++ \cite{yang2020teaser}, Fast Global Registration \cite{zhou2016fast}, and a simple RANSAC-based technique \cite{fischler1981random}. For the global registration algorithms we used FPFH features \cite{rusu2009fast}.  

One important positive characteristic of our benchmark is that, given the large number of problems and their heterogeneity, it is virtually impossible to fine-tune the parameters of an algorithm to each single problem. In this way, the reported results are much more realistic and better represent a real-world usage of the algorithms.

The aim of this section is not to provide a meaningful comparison among the aforementioned approaches. Indeed, the termination criteria, the outlier rejection, and the subsampling methods used are too simple for a real evaluation. Moreover, no effort has been made to fine-tune their parameters.

Instead, our goal is to show how the proposed benchmark should be used, and how sequences coming from different datasets and a large set of initial misalignments and overlaps allow to draw conclusions that would be impossible otherwise.

\subsection{Local Registration Algorithms}

For each algorithm, besides NDT, we used a voxel based subsampling step \cite{rusu2008aligning}, with a leaf-size of $0.1m$ for the TUM datasets and of $0.2m$ for the others. For NDT we used these values as \emph{grid-size}.

ICP and GICP have been tested using the same set of parameters for every sequence, that is:
\begin{itemize}
    \item a uniform subsampling that keeps $70\%$ of the points;
    \item an outlier rejection step based on the median distance. That is, at each step of the algorithm, associations whose distance is greater than three times the median distance of all the associations are discarded \cite{rusu2008aligning};
    \item a termination criterion that stops the algorithm after a maximum of $35$ iterations or when the difference between two consecutive translation estimates is less than $0.01m$, whichever is satisfied earlier.
\end{itemize}

For PPCR, we used the multi-iteration version \cite{fontana2020termination}, using the following parameters:
\begin{itemize}
    \item a uniform subsampling that keeps $30\%$ of the points (to speed up the algorithm);
    \item maximum $20$ neighbours;
    \item an outlier rejection step that keeps only the $70\%$ best (\ie{} closest) correspondences;
    \item a termination criterion that stops the algorithm when the relative cost drop is less than $1\%$ for more than $20$ iterations \cite{fontana2020termination}.
\end{itemize}

Lastly, for NDT we used the same convergence criteria as ICP and G-ICP.

\begin{table}[]
\centering
\small
{\rowcolors{2}{gray!5}{gray!20}
\begin{tabular}{|l|r|r|r|r|r|}
\hline
\textbf{Sequence}       & \textbf{Median} & \textbf{0.75 Q.} & \textbf{0.95 Q.} & \textbf{Mean}             & \textbf{Std Dev} \\ \hline
plain                   & 0.40            & 0.63             & 1.05             & \multicolumn{1}{r|}{0.46} & 0.30             \\
stairs                  & 0.26            & 0.39             & 0.64             & \multicolumn{1}{r|}{0.29} & 0.18             \\
apartment               & 0.36            & 0.55             & 0.98             & \multicolumn{1}{r|}{0.44} & 0.36             \\
hauptgebaude            & 0.13            & 0.25             & 0.56             & \multicolumn{1}{r|}{0.19} & 0.18             \\
wood\_autumn            & 0.34            & 0.43             & 0.57             & \multicolumn{1}{r|}{0.33} & 0.15             \\
wood\_summer            & 0.30            & 0.40             & 0.52             & \multicolumn{1}{r|}{0.30} & 0.14             \\
gazebo\_summer          & 0.24            & 0.35             & 0.52             & \multicolumn{1}{r|}{0.27} & 0.16             \\
gazebo\_winter          & 0.29            & 0.39             & 0.54             & \multicolumn{1}{r|}{0.30} & 0.14             \\ \hline
box\_met                & 0.68            & 1.17             & 2.09             & \multicolumn{1}{r|}{0.85} & 0.63             \\
p2at\_met               & 0.64            & 1.22             & 2.18             & \multicolumn{1}{r|}{0.85} & 0.67             \\
planetary\_map          & 0.66            & 1.26             & 2.21             & \multicolumn{1}{r|}{0.88} & 0.68             \\ \hline
pioneer\_slam           & 0.39            & 0.70             & 3.61             & \multicolumn{1}{r|}{0.90} & 1.85             \\
pioneer\_slam3          & 0.34            & 0.52             & 0.89             & \multicolumn{1}{r|}{0.39} & 0.26             \\
long\_office\_household & 0.59            & 1.10             & 2.26             & \multicolumn{1}{r|}{0.83} & 0.80             \\ \hline
urban05                 & 0.57            & 0.93             & 1.80             & \multicolumn{1}{r|}{0.72} & 0.61             \\ \hline
\rowcolor{green!10}
total                   & 0.36            & 0.60             & 1.58             & \multicolumn{1}{r|}{0.53} & 0.69             \\ \hline
\end{tabular}}
\caption{The results of ICP.\label{tab:results_icp}}
\end{table}

\begin{table}[]
\centering
\small
{\rowcolors{2}{gray!5}{gray!20}
\begin{tabular}{|l|r|r|r|r|r|}
\hline
\textbf{Sequence}       & \textbf{Median} & \textbf{0.75 Q.} & \textbf{0.95 Q.} & \textbf{Mean} & \textbf{Std Dev} \\ \hline
plain                   & 0.12            & 0.29             & 1.27             & 0.29          & 0.48             \\
stairs                  & 0.02            & 0.09             & 0.77             & 0.17          & 0.47             \\
apartment               & 0.07            & 0.78             & 3.24             & 0.70          & 1.24             \\
hauptgebaude            & 0.01            & 0.02             & 0.62             & 0.15          & 0.62             \\
wood\_autumn            & 0.03            & 0.04             & 0.22             & 0.07          & 0.20             \\
wood\_summer            & 0.02            & 0.03             & 0.08             & 0.04          & 0.09             \\
gazebo\_summer          & 0.07            & 0.68             & 2.55             & 0.54          & 0.91             \\
gazebo\_winter          & 0.02            & 0.04             & 0.13             & 0.05          & 0.16             \\ \hline
box\_met                & 2.57            & 5.26             & 9.05             & 3.51          & 3.33             \\
p2at\_met               & 0.55            & 1.60             & 7.09             & 1.58          & 2.83             \\
planetary\_map          & 0.68            & 1.69             & 5.60             & 1.57          & 2.72             \\ \hline
pioneer\_slam           & 0.22            & 0.97             & 5.69             & 1.02          & 1.99             \\
pioneer\_slam3          & 0.11            & 0.25             & 0.68             & 0.21          & 0.29             \\
long\_office\_household & 0.17            & 1.45             & 4.13             & 0.94          & 1.41             \\ \hline
urban05                 & 0.27            & 0.38             & 0.51             & 0.28          & 0.14             \\ \hline
\rowcolor{green!10}
total                   & 0.11            & 0.51             & 3.84             & 0.74          & 1.79             \\ \hline
\end{tabular}}
\caption{The results of G-ICP.\label{tab:results_gicp}}
\end{table}

\Cref{tab:results_gicp,tab:results_icp,tab:results_ndt,tab:results_ppcr} show the results of the four local registration algorithms.

\begin{table}[]
\centering
\small
{\rowcolors{2}{gray!5}{gray!20}
\begin{tabular}{|l|r|r|r|r|r|}
\hline
\textbf{Sequence}       & \textbf{Median} & \textbf{0.75 Q.} & \textbf{0.95 Q.} & \textbf{Mean} & \textbf{Std Dev} \\ \hline
plain                   & 0.65            & 0.89             & 1.18             & 0.63          & 0.35             \\
stairs                  & 0.49            & 0.75             & 1.09             & 0.51          & 0.34             \\
apartment               & 0.45            & 0.72             & 1.03             & 0.46          & 0.34             \\
hauptgebaude            & 0.75            & 1.16             & 1.78             & 0.80          & 0.53             \\
wood\_autumn            & 0.62            & 0.96             & 1.40             & 0.63          & 0.45             \\
wood\_summer            & 0.64            & 0.94             & 1.36             & 0.63          & 0.44             \\
gazebo\_summer          & 0.68            & 1.08             & 1.76             & 0.73          & 0.55             \\
gazebo\_winter          & 0.72            & 1.06             & 1.62             & 0.74          & 0.50             \\ \hline
box\_met                & 1.80            & 2.46             & 3.08             & 1.80          & 0.83             \\
p2at\_met               & 1.72            & 2.42             & 3.06             & 1.74          & 0.85             \\
planetary\_map          & 1.86            & 2.51             & 3.13             & 1.85          & 0.82             \\ \hline
pioneer\_slam           & 0.33            & 0.62             & 1.03             & 0.39          & 0.36             \\
pioneer\_slam3          & 0.23            & 0.62             & 1.15             & 0.38          & 0.45             \\
long\_office\_household & 0.34            & 0.72             & 0.98             & 0.39          & 0.36             \\ \hline
urban05                 & 2.37            & 5.53             & 211.13           & 116.34        & 1104.67          \\ \hline
\rowcolor{green!10}
total                   & 0.72            & 1.25             & 2.89             & 8.53          & 286.49           \\ \hline
\end{tabular}}
\caption{The results of NDT.\label{tab:results_ndt}}
\end{table}

\begin{table}[]
\centering
\small
{\rowcolors{2}{gray!5}{gray!20}
\begin{tabular}{|l|r|r|r|r|r|}
\hline
\textbf{Sequence}       & \textbf{Median} & \textbf{0.75 Q.} & \textbf{0.95 Q.} & \textbf{Mean} & \textbf{Std Dev} \\ \hline
plain                   & 0.17            & 0.46             & 0.93             & 0.30          & 0.35             \\
stairs                  & 0.03            & 0.09             & 0.23             & 0.07          & 0.10             \\
apartment               & 0.07            & 0.28             & 1.50             & 0.34          & 0.68             \\
hauptgebaude            & 0.01            & 0.02             & 0.77             & 0.08          & 0.21             \\
wood\_autumn            & 0.02            & 0.03             & 0.06             & 0.03          & 0.06             \\
wood\_summer            & 0.01            & 0.02             & 0.03             & 0.02          & 0.05             \\
gazebo\_summer          & 0.05            & 0.19             & 0.74             & 0.16          & 0.28             \\
gazebo\_winter          & 0.02            & 0.03             & 0.06             & 0.03          & 0.06             \\ \hline
box\_met                & 1.28            & 2.49             & 4.27             & 1.66          & 1.32             \\
p2at\_met               & 0.47            & 0.97             & 1.96             & 0.68          & 0.63             \\
planetary\_map          & 0.55            & 1.08             & 2.07             & 0.77          & 0.64             \\ \hline
pioneer\_slam           & 0.18            & 0.44             & 3.64             & 0.72          & 1.77             \\
pioneer\_slam3          & 0.15            & 0.34             & 0.75             & 0.25          & 0.31             \\
long\_office\_household & 0.18            & 0.63             & 1.99             & 0.52          & 0.82             \\ \hline
urban05                 & 0.44            & 0.65             & 4.32             & 0.88          & 1.62             \\ \hline
\rowcolor{green!10}
total                   & 0.11            & 0.45             & 1.86             & 0.44          & 0.92             \\ \hline
\end{tabular}}
\caption{The results of PPCR. \label{tab:results_ppcr}}
\end{table}

\begin{figure}
\centering
\includegraphics[width=0.9\linewidth]{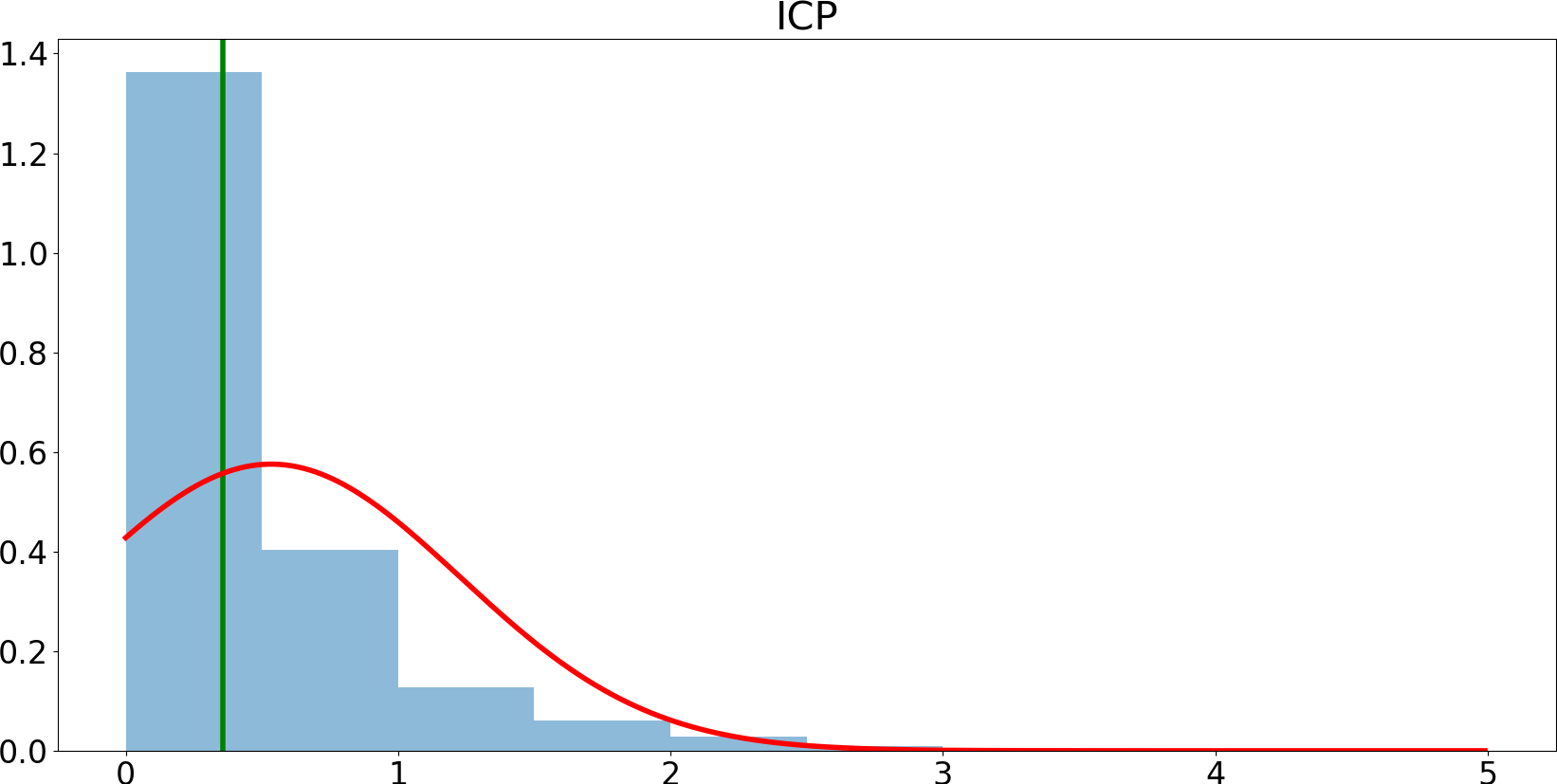}
\par\bigskip
\includegraphics[width=0.9\linewidth]{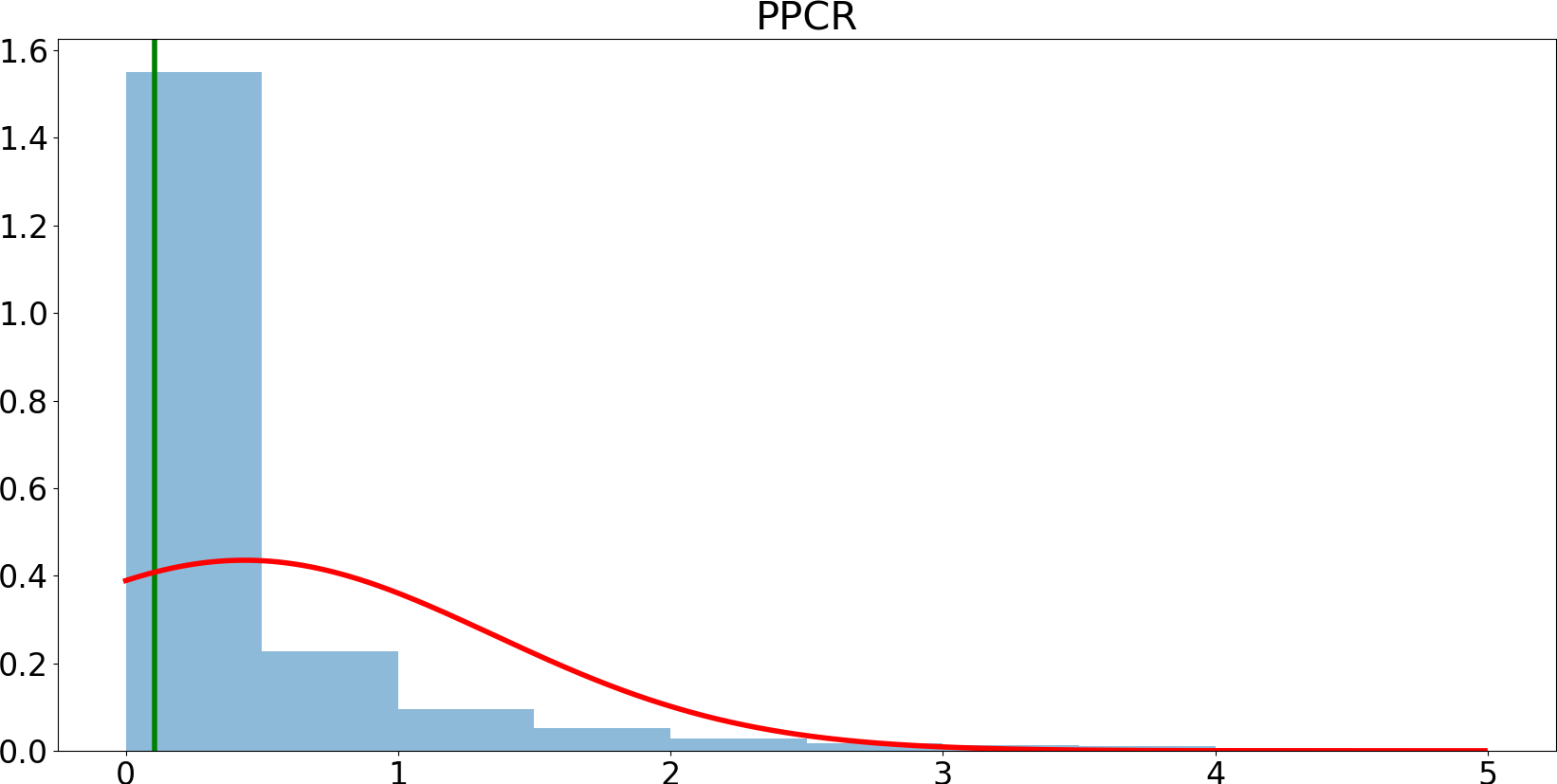}
\caption{Histograms of the results of ICP and PPCR on the whole benchmark. The green line is the median, the red line is a Gaussian built with the mean and standard deviation of the results. As can be seen, the mean does not represent the results properly, since the distribution of the error is far from Gaussian, while the median properly represents the central tendency.\label{fig:hist_1}}
\end{figure}

\begin{figure}
\centering
\includegraphics[width=0.8\linewidth]{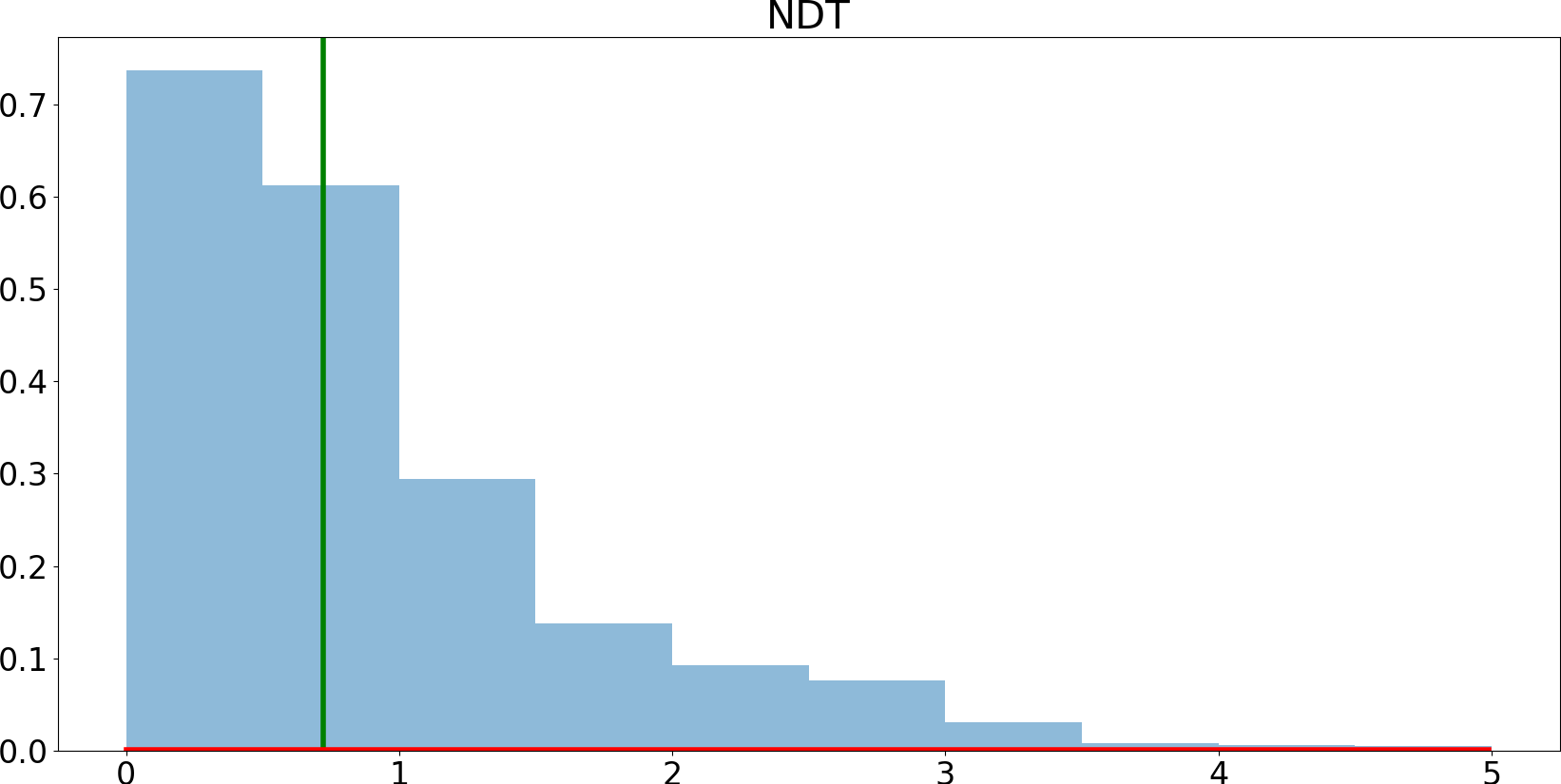}
\par\bigskip
\includegraphics[width=0.8\linewidth]{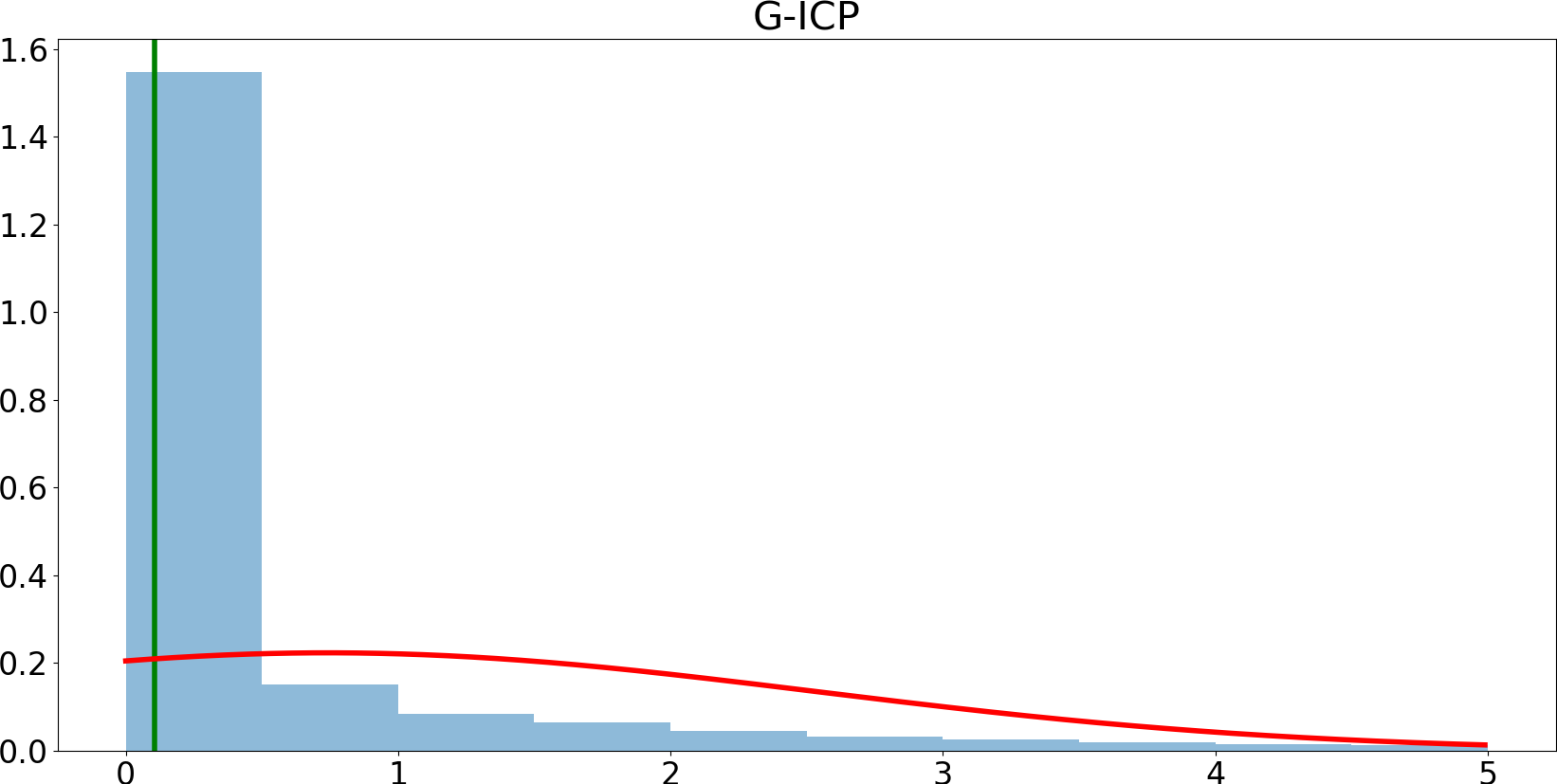}
\caption{Histograms of the results of NDT and G-ICP on the whole benchmark. The green line is the median, the red line is a Gaussian built with the mean and standard deviation of the results. The standard deviation is so large that the Gaussian is almost flat. As can be seen, the mean does not represent the results properly, since the distribution of the error is far from Gaussian, while the median properly represents the central tendency.\label{fig:hist_2}}
\end{figure}

Similarly to Pomerleau \etal{} \cite{pomerleau2013comparing}, we describe the results in terms of quantiles: $0.5$, that is, the median, $0.75$ and $0.95$. Small values of the median mean that the results are accurate, while the precision is evaluated by comparing the difference between different quantiles. If the $0.75$ quantile is very close to the median, it means that $75\%$ of the results are very close to the median, therefore there is a low variance among the results. The same goes for the $0.95$ quantile. Larger differences, on the other hand, mean less precise results, that is, the algorithm finds a good solution less consistently. 

As a measure of the performances of the algorithms, that is the \emph{\textbf{score}} on the benchmark, we use the median of the results. Therefore, the median should be used to objectively compare different algorithms.

The reason behind this choice is that the median describes the results better than the mean and standard deviation, since the error distribution is usually far from Gaussian \cite{pomerleau2013comparing}, see \cref{fig:hist_1,fig:hist_2}. However, we decided to report the mean and the standard deviation too, to compare them to other statistics and show how they do not properly represent the results. As can be seen, for sequences with a large standard deviation the mean is very different from the median; \eg{} in the \emph{pioneer\_slam} sequence, with ICP and PPCR, the mean is more than twice the median. This is even more relevant with the \emph{urban05} sequence for NDT. This phenomenon is due to the median being less affected by extreme values.

We calculated statistics for the whole set of experiments, to compare the four algorithms in a way that we regard as more objective, \ie{}, irregardless of the specific kind of scene. However, since we want to highlight, besides the performances, also the peculiarities of the different approaches, \ie{} in which conditions they perform better and how the structure of the scene affects the result, we calculated statistics also for the single sequences.

\begin{figure}
	\centering
		\includegraphics[width=\linewidth]{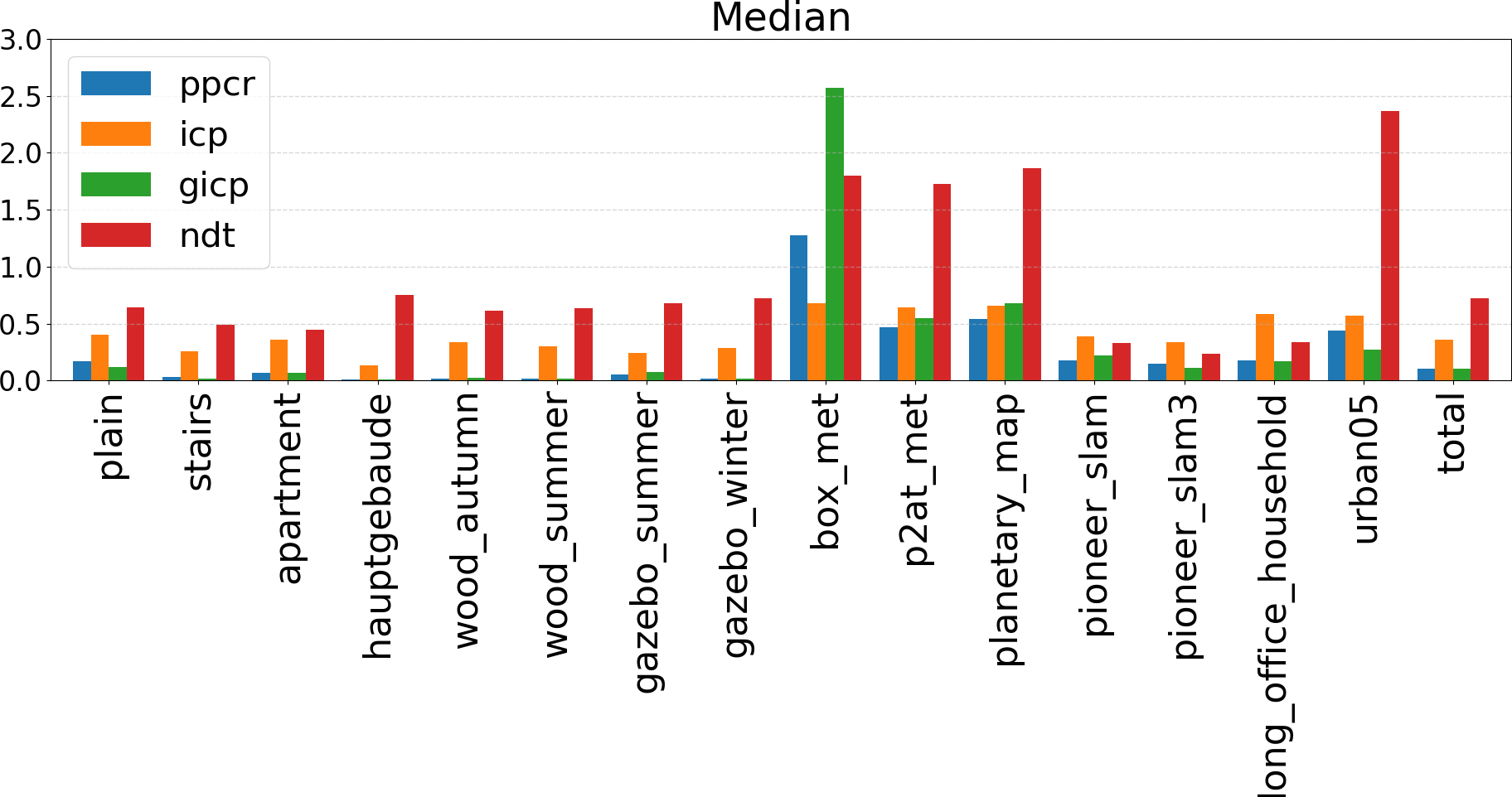}
	\caption{The median results of the local registration algorithms. The \emph{total} column shows the statistics on the whole benchmark, and should be used as an indicator of the performance.\label{fig:results_local}}
\end{figure}

\begin{figure}
	\centering
		\includegraphics[width=\linewidth]{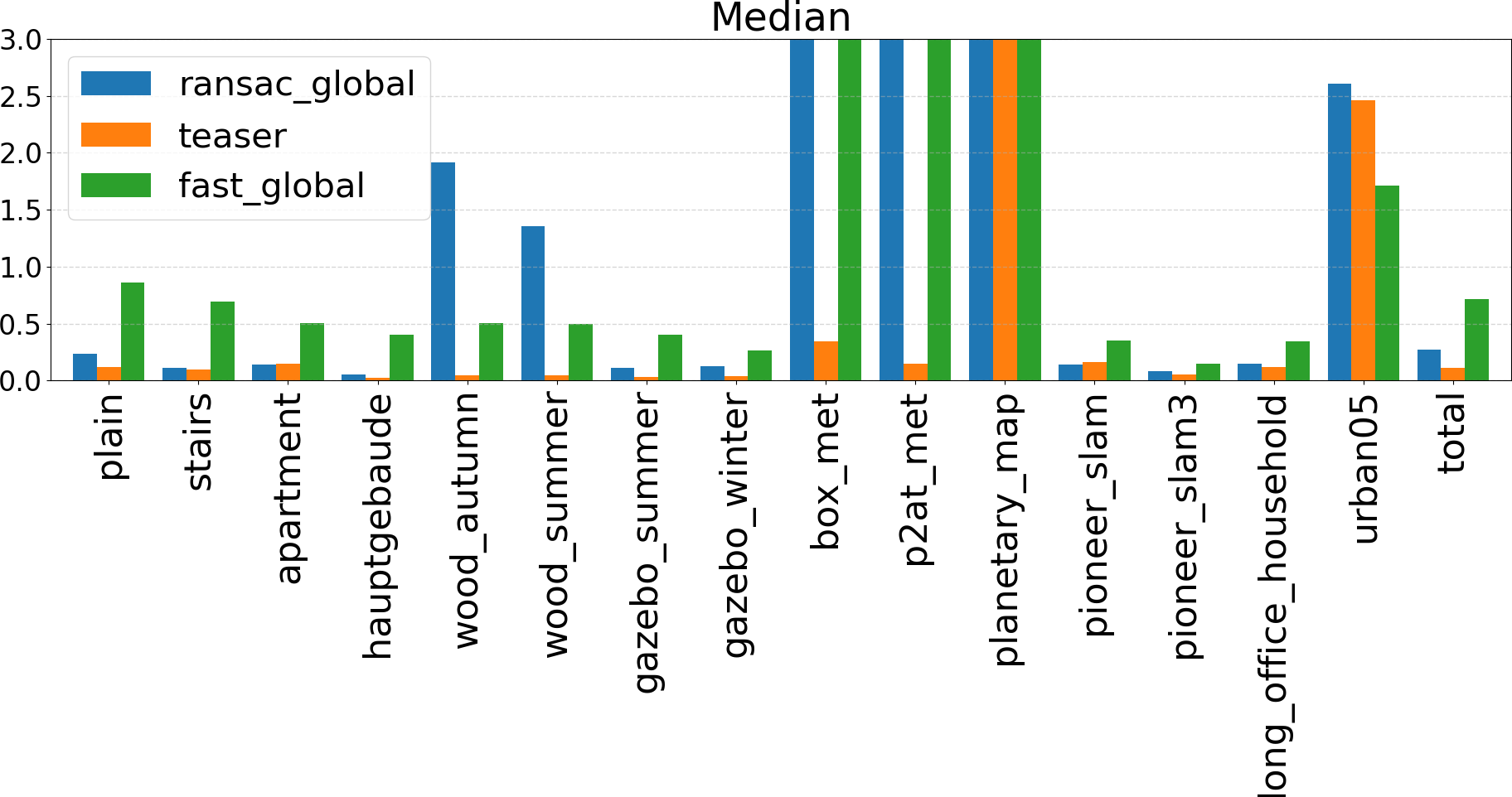}
	\caption{The median results of three algorithms for global registration. The \emph{total} column shows the statistics on the whole benchmark, and should be used as an indicator of the performance. Very large values have been cropped, to better show the smaller differences.\label{fig:results_global}}
\end{figure}

\Cref{fig:results_local} visually represents the median results on different sequences. The last column, named \emph{total}, shows the statistics on the whole benchmark. This last column should be used when comparing the algorithms.

G-ICP and PPCR, in terms of median errors, perform very similarly and much better than ICP and NDT. The only exception is the \emph{box\_met} sequence, where G-ICP got the worst performance. The other algorithms also did not perform well on that sequence. Considering the $0.75$ quantile substantially confirms this conclusion. However, the $0.95$ quantile shows that, when considering the worst cases, PPCR performs better than G-ICP.

It has to be noted that the results among sequences from the same group of datasets (\emph{ETH}, \emph{Planetary}, \emph{TUM} and \emph{Kaist}) are very similar. 
This similarity would not emerge if we were not using many different sequences with different characteristic. Moreover, using only the \emph{ETH} dataset, even though it is a large dataset with many different environments, the performance of the four algorithms would not be correctly estimated, since on that dataset they got a median error that is way smaller than that on the other datasets.

Considering the statistics for the whole benchmark (named ``total'' in the table), we can see how the standard deviation, both for ICP and G-ICP, is so large that the mean does not describe the result properly. This result support the validity of the usage of the median. For example, if we used the mean, we would conclude that ICP would be marginally better than G-ICP. On the contrary, using the median we made the opposite conclusion and highlighted a greater difference between the results.

We also wanted to analyse how the overlap and the initial misalignment affect the result. We expected that a greater overlap leads to an easier registration problem and, therefore, to a better result. On the contrary, a higher initial misalignment should lead to a harder registration problem, especially for local registration algorithms that count on a good initial guess. We wanted to statistically verify this intuition. 

For this reason, for each sequence we calculated the Spearman's rank correlation coefficient between these three variables, \ie{}, overlap, initial misalignment, and the final error on the alignment. The values of the coefficient, along with the p-values, are shown in \cref{tab:corr}. We can observe a correlation \wrt{} the initial misalignment only for NDT (a strong correlation) and for ICP (a moderate correlation). The results of G-ICP and PPCR, instead, do not appear to heavily depend on the initial misalignment. Moreover, there is no correlation \wrt{} the overlap. All but one of the p-values are very small, so we can have a high confidence in the reported correlation coefficients.

It appears that the result is much more affected by the structure of the scene than by the overlap and the initial misalignment. This conclusion highlights the importance of uniformly sampling a vast range of overlaps and initial misalignments, as we did with our benchmark. It is only thanks to this characteristic that we could objectively analyze the behaviour of the algorithms in many different conditions and draw these counter-intuitive conclusions. This statement can be  supported only because of the heterogeneity of the testing conditions. Of course this conclusion is valid only for the tested algorithms (including the subsampling and outlier rejection methods) and for the specific set of parameters used. 

\begin{table}[]
\centering
\small
\begin{tabular}{|l||c|c|}
\hline
\textbf{Algorithm} & \textbf{\begin{tabular}[c]{@{}l@{}}Corr. \\ Initial Misalignment \\ (p-value)\end{tabular}} & \textbf{\begin{tabular}[c]{@{}l@{}}Corr. \\ Overlap (p-value)\end{tabular}} \\ \hline \hline
\rowcolor[HTML]{EFE9E9} 
ICP                & 0.6940 (\textless{}0.001)                                                                   & 0.0013 (0.7810)                                                             \\
G-ICP              & 0.4088 (\textless{}0.001)                                                                   & -0.2347 (\textless{}0.001)                                                  \\
\rowcolor[HTML]{EFE9E9} 
PPCR               & 0.4683 (\textless{}0.001)                                                                   & -0.1534 (\textless{}0.001)                                                  \\
NDT                & 0.8743 (\textless{}0.001)                                                                   & -0.1033 (\textless{}0.001)                                                  \\
\rowcolor[HTML]{EFE9E9} 
Fast Global        & 0.2290 (\textless{}0.001)                                                                   & -0.2442 (\textless{}0.001)                                                  \\
Teaser             & 0.3208 (\textless{}0.001)                                                                   & -0.1588 (\textless{}0.001)                                                  \\
\rowcolor[HTML]{EFE9E9} 
RANSAC             & 0.1549 (\textless{}0.001)                                                                   & -0.2310 (\textless{}0.001)                                                  \\ \hline
\end{tabular}
\caption{The Spearman's correlation coefficient between the performance measure and the initial misalignment and overlap. In brackets the corresponding p-value. \label{tab:corr}}
\end{table}

\subsection{Global Registration Algorithms}
For the global registration algorithms (RANSAC, Teaser and Fast Global) we used the same sub-sampling method as the one employed for the local registration algorithms, with the same voxel-size values. 

To compute the FPFH feature descriptors we used a radius of five times the voxel size, while, to estimate the normals needed by the descriptors, we used a radius of twice the voxel size \cite{rusu2009fast}.

\Cref{tab:results_fast_global,tab:results_ransac,tab:results_teaser} show the results of the algorithms on the benchmark, while \cref{fig:results_global} visually compares the medians. Teaser obtained excellent results on most of the sequences, with performances comparable to those of local registration algorithms (on the set of local registration problems). It was even able to properly align point clouds from the \emph{box\_met} and \emph{p2at\_met} sequences, where RANSAC and Fast Global failed catastrophically. However, it was not able to obtain acceptable results on the \emph{planetary\_map} and \emph{urban05} sequences (but neither were the other algorithms). The failure on the \emph{planetary\_map} sequence is probably due to the kind of feature used, more than to the algorithms. The map used in that sequence, indeed, is relatively sparse; therefore, using features based on the normal to the surface becomes problematic. Moreover, the surface is highly repetitive, hence the features are less discriminative. 

The results of Fast Global and RANSAC heavily depend on the dataset, although neither of the two performed properly on the Planetary and Kaist datasets. If the benchmark was not composed of many different datasets, the evaluation of the algorithms would be highly biased, since, as it can be seen, the results heavily depend on the sequence used (especially for RANSAC and Fast Global).

The experiments with the global registration algorithms properly show why we used the median instead of the mean. Looking at the mean, Teaser would be considered the worst of the three algorithms. This is because of its very poor performance on a single sequence (\emph{planetary\_map}). However, on that sequence none of the algorithms was able to obtain acceptable results. Therefore, even tough the numerical difference between the results is large, it is definitely not relevant. On the other hand, Teaser performed very well on most of the other sequences and this is reflected by its very low median result.

\Cref{tab:corr} shows that, as expected, the performances of the global algorithms are not correlated with the initial misalignment. This is true also for the overlap, which is not an expected behaviour but is consistent with what observed for the local registration algorithms.
\begin{table}[]
\centering
\small
{\rowcolors{2}{gray!5}{gray!20}
\begin{tabular}{|l|r|r|r|r|r|}
\hline
\textbf{Sequence}       & \textbf{Median} & \textbf{0.75 Q.} & \textbf{0.95 Q.} & \textbf{Mean} & \textbf{Std Dev} \\ \hline
plain                   & 0.12            & 0.29          & 2.53          & 0.60          & 1.61             \\
stairs                  & 0.09            & 0.44          & 4.14          & 0.79          & 1.35             \\
apartment               & 0.15            & 1.54          & 3.25          & 0.94          & 1.80             \\
hauptgebaude            & 0.03            & 1.01          & 2.04          & 0.56          & 0.73             \\
wood\_autumn            & 0.05            & 0.08          & 0.25          & 0.08          & 0.12             \\
wood\_summer            & 0.05            & 0.08          & 0.23          & 0.08          & 0.13             \\
gazebo\_summer          & 0.03            & 0.07          & 0.47          & 0.14          & 0.43             \\
gazebo\_winter          & 0.04            & 0.06          & 0.40          & 0.12          & 0.29             \\ \hline
box\_met                & 0.34            & 8.59          & 13.12         & 3.86          & 5.15             \\
p2at\_met               & 0.15            & 0.88          & 10.99         & 1.91          & 3.61             \\
planetary\_map          & 44.54           & 64.74         & 95.66         & 49.97         & 24.57            \\ \hline
pioneer\_slam           & 0.16            & 0.35          & 2.12          & 0.64          & 1.45             \\
pioneer\_slam3          & 0.06            & 0.12          & 1.01          & 0.28          & 1.01             \\
long\_office\_household & 0.12            & 0.45          & 2.76          & 0.91          & 2.42             \\ \hline
urban05                 & 2.46            & 3.87          & 7.62          & 3.04          & 2.23             \\ \hline
\rowcolor{green!10}
total                   & 0.11            & 1.16          & 34.04         & 4.32          & 14.07            \\ \hline
\end{tabular}}
\caption{The results of TEASER++ \label{tab:results_teaser}}

\end{table}

\begin{table}[]
\centering
\small
{\rowcolors{2}{gray!5}{gray!20}
\begin{tabular}{|l|r|r|r|r|r|}
\hline
\textbf{Sequence}       & \textbf{Median} & \textbf{0.75 Q.} & \textbf{0.95 Q.} & \textbf{Mean} & \textbf{Std Dev} \\ \hline
plain                   & 0.86            & 1.56          & 2.49          & 1.13          & 0.98             \\
stairs                  & 0.69            & 1.78          & 4.06          & 1.19          & 1.33             \\
apartment               & 0.50            & 1.72          & 2.98          & 1.05          & 1.11             \\
hauptgebaude            & 0.40            & 0.99          & 1.77          & 0.60          & 0.54             \\
wood\_autumn            & 0.50            & 0.79          & 2.01          & 0.67          & 0.55             \\
wood\_summer            & 0.50            & 0.79          & 1.71          & 0.63          & 0.48             \\
gazebo\_summer          & 0.40            & 0.88          & 1.92          & 0.64          & 0.72             \\
gazebo\_winter          & 0.27            & 0.65          & 1.44          & 0.49          & 0.50             \\ \hline
box\_met                & 7.24            & 9.32          & 13.66         & 7.58          & 3.21             \\
p2at\_met               & 3.37            & 6.27          & 9.25          & 4.03          & 3.03             \\
planetary\_map          & 29.58           & 42.71         & 67.71         & 32.37         & 17.91            \\ \hline
pioneer\_slam           & 0.35            & 1.63          & 10.42         & 2.20          & 3.62             \\
pioneer\_slam3          & 0.15            & 0.27          & 0.98          & 0.44          & 1.18             \\
long\_office\_household & 0.35            & 3.11          & 11.09         & 2.38          & 3.65             \\ \hline
urban05                 & 1.71            & 2.17          & 3.46          & 1.74          & 0.98             \\ \hline
\rowcolor{green!10}
total                   & 0.71            & 2.51          & 18.15         & 3.85          & 9.38             \\ \hline
\end{tabular}}
\caption{The results of Fast Global Registration \label{tab:results_fast_global}}
\end{table}

\begin{table}[]
\centering
\small
{\rowcolors{2}{gray!5}{gray!20}
\begin{tabular}{|l|r|r|r|r|r|}
\hline
\textbf{Sequence}       & \textbf{Median} & \textbf{0.75 Q.} & \textbf{0.95 Q.} & \textbf{Mean} & \textbf{Std Dev} \\ \hline
plain                   & 0.23            & 0.50          & 3.29          & 0.63          & 1.15             \\
stairs                  & 0.11            & 0.20          & 3.14          & 0.58          & 1.27             \\
apartment               & 0.14            & 0.22          & 1.74          & 0.35          & 0.76             \\
hauptgebaude            & 0.05            & 0.74          & 2.51          & 0.49          & 0.83             \\
wood\_autumn            & 1.92            & 3.08          & 4.90          & 1.97          & 1.60             \\
wood\_summer            & 1.36            & 2.39          & 4.04          & 1.50          & 1.46             \\
gazebo\_summer          & 0.11            & 0.96          & 3.71          & 0.80          & 1.33             \\
gazebo\_winter          & 0.13            & 0.77          & 3.90          & 0.79          & 1.29             \\ \hline
box\_met                & 7.58            & 10.38         & 13.61         & 7.64          & 3.72             \\
p2at\_met               & 4.85            & 9.59          & 15.05         & 5.73          & 5.59             \\
planetary\_map          & 10.80           & 15.11         & 20.78         & 11.65         & 5.13             \\ \hline
pioneer\_slam           & 0.14            & 0.35          & 1.67          & 0.63          & 1.94             \\
pioneer\_slam3          & 0.09            & 0.14          & 0.40          & 0.28          & 1.18             \\
long\_office\_household & 0.15            & 0.24          & 1.96          & 0.78          & 2.79             \\ \hline
urban05                 & 2.61            & 3.71          & 5.30          & 2.87          & 1.38             \\ \hline
\rowcolor{green!10}
total                   & 0.27            & 3.01          & 12.10         & 2.46          & 4.13             \\ \hline
\end{tabular}}
\caption{The results of RANSAC \label{tab:results_ransac}}
\end{table}

\section{Conclusions}
In the field of point clouds registration, most approaches are tested on very few data, often collected ad-hoc. For this reason, the results are hardly generalizable and do not allow a comparison with other approaches tested on different data. We present a new benchmark for point clouds registration algorithms, whose main goal is to allow a rigorous comparison between different approaches. It is composed of several publicly available datasets, chosen to cover an extensive set of scenarios and use cases, including settings usually neglected, but still very relevant for some real applications. The benchmark can be used to test both global and local registration algorithms, with different initial misalignments and different degrees of overlap. For each sequence, in each dataset, we randomly chose a list of pairs of point clouds in a way that ensures uniform coverage of the various degrees of overlap. For each pair, we randomly generated a list of rototranslations to be applied as initial misplacements. These rotoranslations are, indeed, the transformations that a registration algorithm should estimate.

We also propose a new metric to measure the residual error after the registration and to objectively compare different approaches. To encourage the use of our benchmark, we developed a set of utilities to setup the testing environment and to calculate the necessary metrics: our goal was to reduce the effort required to use the benchmark as much as possible.

Instructions on how to use the protocol and the related utilities are available on the project web page: \url{https://github.com/iralabdisco/point_clouds_registration_benchmark}. We would be glad to list works that use our benchmark. To add a paper to the list, please contact the corresponding author.





\bibliographystyle{elsarticle-num-names}
\bibliography{biblio.bib}







\end{document}